\pdfoutput=1
\documentclass[10pt,letterpaper]{article}

\usepackage[margin=1in]{geometry}
\usepackage[utf8]{inputenc}
\usepackage[T1]{fontenc}
\usepackage{amsmath,amssymb}
\usepackage{booktabs}
\usepackage{tabularx}
\usepackage[numbers,sort&compress]{natbib}
\usepackage{hyperref}

\usepackage{llamapaper}
\usepackage[nameinlink,capitalize]{cleveref}
\usepackage{pgfplots}
\pgfplotsset{compat=1.18}
\usepackage{placeins}
\usepackage{comment}

\def\cleanbuildenabled{}
\ifdefined\cleanbuildenabled
\newcommand{\simon}[1]{}
\newcommand{\sebas}[1]{}
\newcommand{\nutch}[1]{}
\newcommand{\antonio}[1]{}
\newcommand{\pierre}[1]{}
\else
\newcommand{\simon}[1]{{\color{blue}\textbf{[Simon:} #1\textbf{]}}}
\newcommand{\sebas}[1]{{\color{orange!85!black}\textbf{[Sebas:} #1\textbf{]}}}
\newcommand{\nutch}[1]{{\color{red!70!black}\textbf{[Nutch:} #1\textbf{]}}}
\newcommand{\antonio}[1]{{\color{green}\textbf{[Antonio:} #1\textbf{]}}}
\newcommand{\pierre}[1]{{\color{violet!75!black}\textbf{[Pierre:} #1\textbf{]}}}
\fi

\newcommand{\ocria}{OCRia-1}

\newcommand{\coremark}{\textsuperscript{*}}
\newcommand{\core}[1]{#1\coremark}
\newcommand{\corresp}{\textsuperscript{\textdagger}}

\newcommand{\RCsetup}{\centering\footnotesize\setlength{\tabcolsep}{5pt}%
  \setlength{\aboverulesep}{0pt}\setlength{\belowrulesep}{0pt}}
\newcommand{\RChead}[1]{\rule[-1.1ex]{0pt}{3.4ex}{\sffamily\bfseries #1}}
\newcommand{\RCgroup}[2]{\noalign{\vspace{1pt}\hbox{\setlength{\fboxsep}{5pt}\colorbox{LIPanelBar}{%
  \makebox[\dimexpr\linewidth-2\fboxsep][l]{\textcolor{LIBlue}{\rule{4pt}{4pt}}\hspace{0.5em}%
  {\scriptsize\sffamily\bfseries\color{LIMuted}\MakeUppercase{#1}}\hfill
  {\scriptsize\color{LIMuted}#2}}}}\vspace{4pt}}}

\newcommand{\RCnext}{\\\RCsep}
\newcommand{\RClast}{\\\noalign{\vspace{3.5pt}}}
\newcommand{\RCstart}{\noalign{\vspace{3.5pt}}}

\title{OCRia-1 Technical Report}
\author{
  \core{Aman Rangapur} \quad \core{Antonio Jimeno Yepes}\corresp \quad \core{Daniel Bustamante} \\
  Nutchapol Dendumrongsup \quad Sebastián G. Acosta \quad Pierre-Loïc Doulcet \quad Simon Suo \\[4pt]
  {\mdseries\footnotesize\color{LIMuted}\texttt{amanrangapur@gmail.com}, \texttt{\{antonio, daniel, simon\}@runllama.ai}\\
   \coremark\ core contributors\enspace\textperiodcentered\enspace\textsuperscript{\textdagger}\,corresponding author}
}
\date{}

\LIPaperDate{October 2026}
\LIPDFAuthor{Aman Rangapur, Antonio Jimeno Yepes, Daniel Bustamante}
\LIPDFSubject{OCR for document parsing}
\LIAbstract{
We introduce \textbf{\ocria}, a 4B vision-language model (VLM) for document parsing that excels at preserving table structure, extracting chart data and retaining text formatting.
\ocria{} handles content that has traditionally been challenging in document parsing: merged cells and spanning headers, tables with thousands of cells, chart data,
and meaningful text formatting such as bold, italics, subscripts and superscripts, preserving details needed for
business-critical document analysis.
At the
time of this report, it achieves the highest scores on tables, charts and text formatting among the systems evaluated, including
a 35B model and Mistral OCR 4.1. Its mean ParseBench score over the four dimensions we evaluate is 0.87, against 0.75 for the next-best system.

\smallskip
We emphasize data quality and reward design in training \ocria{}. Synthetic pages provide exact ground truth for
long tables and rich formatting, while real pages are transcribed using a pipeline grounded in the PDF text layer.
We attribute \ocria{}'s performance primarily to careful data curation that prioritizes quality and diversity over quantity.
Dimension-specific rewards evaluate table reconstruction, chart-data extraction and text-formatting
fidelity. We train the model on the curated data using supervised fine-tuning (SFT), then refine it through
reinforcement learning with verifiable rewards (RLVR).

\smallskip

To our knowledge, \ocria{} is the fastest published document-parsing VLM in its class, decoding at about 4,700 tokens per second with a single request on one B200 GPU, about nine times faster than the base model on stock vLLM. This is achieved through a block-diffusion speculator, quantization and custom kernels.
}
\LIHero{%
  {\centering\definecolor{HRvermillion}{HTML}{D55E00}\definecolor{HRgreen}{HTML}{009E73}%
\definecolor{HRorange}{HTML}{E69F00}\definecolor{HRpurple}{HTML}{CC79A7}%
\begin{tikzpicture}[font=\sffamily,
    hrO/.style={LIBlue, line width=1.3pt},
    hrA/.style={HRvermillion, line width=1.0pt},
    hrB/.style={HRgreen, line width=1.0pt},
    hrC/.style={HRorange, line width=1.0pt},
    hrD/.style={HRpurple, line width=1.0pt}]
  \def\HRr{3.1}
  \def\HRmin{30}
  \def\HRpoly#1#2{%
    \foreach \v [count=\i] in {#2} {\coordinate (HRp\i) at ({90-360/6*(\i-1)}:{(\v-\HRmin)/(100-\HRmin)*\HRr});}
    \draw[#1] (HRp1)--(HRp2)--(HRp3)--(HRp4)--(HRp5)--(HRp6)--cycle;}
  \foreach \g in {40,60,80,100} {
    \foreach \k in {1,...,6} {\coordinate (HRg\k) at ({90-360/6*(\k-1)}:{(\g-\HRmin)/(100-\HRmin)*\HRr});}
    \draw[LIPanelLine, line width=0.3pt] (HRg1)--(HRg2)--(HRg3)--(HRg4)--(HRg5)--(HRg6)--cycle;
    \node[font=\fontsize{6.5pt}{6.5pt}\selectfont\sffamily, text=LIMuted, anchor=east, inner sep=1pt, xshift=-1.5pt, fill=white, fill opacity=0.85, text opacity=1]
      at ({90}:{(\g-\HRmin)/(100-\HRmin)*\HRr}) {\g};
  }
  \foreach \k in {0,...,5} {\draw[LIPanelLine, line width=0.3pt] (0,0) -- ({90-360/6*\k}:\HRr);}
  \foreach \k/\anch/\lab in {
      0/south/{\textbf{Tables}\\[-1pt]{\color{LIMuted}ParseBench}},
      1/south west/{\textbf{Charts}\\[-1pt]{\color{LIMuted}ParseBench}},
      2/north west/{\textbf{Text formatting}\\[-1pt]{\color{LIMuted}ParseBench}},
      3/north/{\textbf{Text content}\\[-1pt]{\color{LIMuted}ParseBench}},
      4/north east/{\textbf{olmOCR-bench}\\[-1pt]{\color{LIMuted}tables and formulas}},
      5/south east/{\textbf{PulseBench-Tab}\\[-1pt]{\color{LIMuted}tables, nine languages}}} {
    \node[anchor=\anch, align=center, font=\footnotesize\sffamily, text=LIInk, inner sep=2pt]
      at ({90-360/6*\k}:{\HRr+0.12}) {\lab};
  }
  \HRpoly{hrA}{89,65,61,84,89.4,80}
  \HRpoly{hrB}{83,66,64,89,90.1,82}
  \HRpoly{hrC}{86,61,59,90,90.2,82}
  \HRpoly{hrD}{74,40,66,90,80.3,75}
  \HRpoly{hrO, fill=LIBlue, fill opacity=0.10}{92,85,83,90,90.5,84}
  \foreach \i in {1,...,6} {\fill[LIBlue] (HRp\i) circle (1.3pt);}
  \begin{scope}[shift={(\HRr+3.3,1.9)}]
    \foreach \y/\sty/\lab in {
        0/hrO/{\textbf{\ocria{} (4B)}},
        -0.5/hrA/{Chandra-OCR-2 (4B)},
        -1.0/hrB/{LightOnOCR-3-4B},
        -1.5/hrC/{Infinity-Parser2-Pro (35B)},
        -2.0/hrD/{Mistral OCR 4.1}} {
      \draw[\sty] (0,\y) -- (0.65,\y);
      \node[anchor=west, font=\footnotesize\sffamily, text=LIInk, inner sep=2pt] at (0.7,\y) {\lab};
    }
  \end{scope}
\end{tikzpicture}\par}
  \captionof{figure}{\ocria{} has the highest score on every axis against leading document-parsing VLMs, on the four ParseBench dimensions and on olmOCR-bench and PulseBench-Tab from \Cref{tab:main-results}, with olmOCR-bench restricted to its table and formula categories. Scores are on a 0--100 scale, and the radial axis starts at 30.}
  \label{fig:hero}
}

\begin{document}

\maketitle

\section{Introduction}
\label{introduction}

Document parsing converts a document page into text that keeps the structure of the page, such as its headings, reading order, tables and formulas.
Some documents store their text but not its structure; a born-digital PDF, for example, records characters and their positions, but not which characters form a heading or a paragraph~\citep{poznanski2025olmocrunlockingtrillionstokens}.
Others, such as scanned or photographed pages, are only images.
In both cases, tables, charts, formatting and formulas are among the hardest elements to recover.
Parsing mistakes are costly, because retrieval systems, extraction pipelines and document agents read only the parsed text and cannot correct it~\citep{zhang2024documentparsingunveiled,zhang2026parsebenchdocumentparsingbenchmark,zhang2024ocrhindersrag}.
For example, a value placed under the wrong column header becomes a wrong answer, although the table still looks well formed~\citep{zhang2026parsebenchdocumentparsingbenchmark}.
Parsing must also be fast, because training-data generation and retrieval indexing process millions of pages, and document agents need low-latency parsing to answer questions quickly.

Compact document-parsing VLMs have become the standard way to parse documents, and a 1--4B parameter model that reads the page now replaces the heuristic pipelines in most open systems~\citep{li2025dotsocrmultilingualdocumentlayout,taghadouini2026lightonocr1bendtoendmultilingual,lightonocr3blog2026,duan2026glmocrtechnicalreport,cai2026teleocrnavigatingdocumentparsing}.
These models read plain text well.
They are less accurate on complex tables and charts.
Tables with merged cells, spanned headers or irregular shapes come out with cells in the wrong place, and long tables are often cut short or lose rows.
Charts are usually treated as pictures, so most models describe them or skip them instead of writing out their data.
These models are also slow.
They write one token at a time, and a dense page of tables runs to thousands of tokens, so parsing thousands of pages takes hours of GPU time.
We built \ocria{} for these elements and for speed.

We introduce \textbf{\ocria}, a 4B document-parsing VLM built on Qwen3.5-4B~\citep{qwen3.5,qwen3.5-4b}.
\ocria{} transcribes the page into Markdown, writing tables as HTML with their full structure, charts as tables of their data points, and formulas as \LaTeX.
\ocria{} is more accurate on tables, charts and text formatting than every system we compare it with (\Cref{fig:hero}), and its mean ParseBench score over the four dimensions we evaluate is 0.87, against 0.75 for the next-best system (\Cref{tab:main-results}).
It is also, to our knowledge, the fastest published document-parsing VLM.
On one B200 GPU, the served model decodes at about 4,700 tokens per second with a single request and about 1,900 tokens per second per request with eight concurrent requests.
With 64 concurrent requests, it parses about 10 pages per second, at no measurable cost on the benchmarks (\Cref{eval:speed}).
We plan to make a version of \ocria{} available.

We gave priority to tables, charts and text formatting in both the data and the training.
These elements carry much of the information in enterprise documents, such as financial reports, insurance filings, contracts and medical publications.
Our approach is to invest in training data for these elements, in rewards that score them directly, and in inference speed.
The training set has two kinds of pages (\Cref{data}).
Real pages come from public document collections, chosen where document-parsing VLMs still fail, such as dense tables, tables with cells that span several rows or columns, and charts.
A pipeline of proprietary vision-language models transcribes each real page, and we keep the transcription only when independent checks agree with it (\Cref{annotations}).
Synthetic pages come from our own generators, which render varied page layouts together with their exact transcription.
They supply what real collections do not contain in quantity, such as long tables, icon tables and rich inline formatting.
Supervised fine-tuning (SFT) teaches the model to transcribe these pages, and reinforcement learning with verifiable rewards (RLVR) then rewards it for table cells placed under the right headers, chart values and formatting marks.

To accelerate decoding, we use speculative decoding~\citep{leviathan2023fast,chen2023accelerating} with a block-diffusion draft model~\citep{dflash2026}, and adapt the number of draft tokens to the batch size to maintain high throughput at larger batch sizes.
We also quantized \ocria{} to four-bit NVFP4, while retaining FP8 or BF16 in layers where lower-precision quantization reduced evaluation scores.
Finally, we customized vLLM with architecture-specific CUDA kernels and fused operations to improve GPU compute and memory-bandwidth utilization.

This report provides more information about how we trained \ocria{} and is organized as follows.
\Cref{data} describes our diverse, high-quality mix of real and synthetic data, which we view as the primary driver of \ocria{}'s performance.
\Cref{training} describes SFT and RLVR with rewards that score table cells and chart values.
\Cref{inference} describes speculative decoding, quantization and serving optimizations.
\Cref{evaluations} compares \ocria{} with eight other systems on four transcription benchmarks and measures its inference speed on one B200 GPU.
\Cref{related} places this work among other document-parsing VLMs.

\section{Data}
\label{data}

Our data is collected strategically around the most challenging aspects of document parsing, including charts and long tables or tables with many spanning cells (\cref{related}).
We developed a proprietary workflow to build a high-quality training corpus from both transcribed public documents and synthesised pages, to close the gaps that remain in public datasets.
We used synthetic data to fill the gaps left by real data, which ensures the variety of hard cases needed for training.
This allows us to include the long tail of difficult cases without the struggle of finding them in real data.
Synthetic data comes with training ground truth by design, and we carefully designed the layouts of the synthetic examples to be as realistic as possible.
This strategy ensures that the training set covers the aspects a model needs to get right, while keeping our customer data outside the training corpus.
\Cref{corpus} describes what we collected and why.
\Cref{annotations} describes how we generate ground truth for a real page: a teacher pipeline writes it, and independent checks decide whether we keep it.
\Cref{synthetic} describes how the synthetic pages are generated.

\begin{figure}[t]
\centering
\begingroup
\colorlet{DOReal}{LIBlue!85!black}
\colorlet{DOSynthetic}{LICyan!42!black}
\begin{tikzpicture}[x=1cm, y=1cm,
    >={Stealth[length=4.5pt,width=3.5pt]},
    every node/.style={text=LIInk, align=center,
      font=\fontsize{8pt}{10pt}\selectfont\sffamily},
    lane label/.style={anchor=north west, inner sep=0pt,
      font=\fontsize{8pt}{9pt}\selectfont\sffamily\bfseries},
    stage/.style={draw=LIPanelLine, fill=white, line width=0.5pt,
      rounded corners=4pt, minimum width=2.9cm, minimum height=1.4cm,
      text width=2.55cm, inner xsep=4pt, inner ysep=6pt,
      execute at begin node={\hyphenpenalty=10000\relax}},
    product/.style={draw=DOSynthetic!30!white, fill=white,
      line width=0.5pt, rounded corners=4pt,
      minimum width=3.25cm, minimum height=0.72cm,
      inner sep=4pt, text=DOSynthetic,
      font=\fontsize{8.5pt}{10pt}\selectfont\sffamily\bfseries},
    real flow/.style={->, draw=DOReal, line width=0.8pt, rounded corners=4pt},
    synthetic flow/.style={->, draw=DOSynthetic, line width=0.8pt, rounded corners=4pt}]

  \fill[LIBlue!5!white, rounded corners=6pt] (0,3.65) rectangle (11.15,6.3);
  \fill[LICyan!7!white, rounded corners=6pt] (0,0) rectangle (11.15,3.4);
  \node[lane label, text=DOReal] at (0.3,6.02) {REAL PAGES};
  \node[lane label, text=DOSynthetic] at (0.3,3.12) {SYNTHETIC PAGES};
  \node[lane label, anchor=north east, text=DOReal]
    at (10.85,6.02) {Real-world diversity};
  \node[lane label, anchor=north east, text=DOSynthetic]
    at (10.85,3.12) {Hard cases with exact targets};

  \node[stage, minimum width=3.4cm, text width=3.1cm] (real) at (1.95,4.9)
    {{\bfseries Collect real pages}\\[5pt]Languages, layouts,\\scans, handwriting};
  \node[stage] (transcribe) at (5.45,4.9)
    {{\bfseries Transcribe}\\[5pt]Read tables,\\charts and text};
  \node[stage, minimum width=3.25cm, text width=2.95cm] (check) at (8.95,4.9)
    {{\bfseries Independent checks}\\[5pt]Image + PDF text;\\correct errors};
  \draw[real flow] (real.east) -- (transcribe.west);
  \draw[real flow] (transcribe.east) -- (check.west);

  \node[stage, minimum width=3.4cm, text width=3.1cm] (content) at (1.95,2.0)
    {{\bfseries Generate hard cases}\\[5pt]Long tables and\\spanned headers};
  \node[stage] (render) at (5.45,2.0)
    {{\bfseries Render pages}\\[5pt]Styles from\\real documents};
  \node[product] (page) at (8.95,2.0) {Page image};
  \node[product] (target) at (8.95,0.65) {Exact transcription};
  \draw[synthetic flow] (content.east) -- (render.west);
  \draw[synthetic flow] (render.east) -- (page.west);
  \draw[synthetic flow] (content.south) -- (1.95,0.65) -- (target.west);

  \draw[draw=LIPanelLine, fill=white, line width=0.7pt, rounded corners=5pt]
    (12.4,1.95) rectangle (16.3,4.4);
  \node[font=\fontsize{10pt}{12pt}\selectfont\sffamily\bfseries]
    at (14.35,4.02) {Training pairs};
  \draw[LIPanelLine, line width=0.5pt] (12.65,3.72) -- (16.05,3.72);

  \draw[draw=LIFlowGray!65!white, fill=LIPanelGray, line width=0.6pt,
    rounded corners=1pt] (13.0,2.65) rectangle (13.8,3.5);
  \fill[LIFlowGray!65!white] (13.12,3.29) rectangle (13.68,3.34);
  \draw[draw=LIFlowGray!65!white, line width=0.4pt]
    (13.12,2.8) rectangle (13.68,3.16);
  \draw[draw=LIFlowGray!65!white, line width=0.4pt]
    (13.12,2.92) -- (13.68,2.92)
    (13.12,3.04) -- (13.68,3.04)
    (13.3,2.8) -- (13.3,3.16)
    (13.49,2.8) -- (13.49,3.16);
  \node[font=\fontsize{14pt}{14pt}\selectfont\sffamily, text=LIMuted]
    at (14.35,3.08) {+};
  \draw[draw=LIFlowGray!65!white, fill=LIPanelGray, line width=0.6pt,
    rounded corners=1pt] (14.9,2.65) rectangle (15.7,3.5);
  \foreach \y/\w in {3.3/0.56,3.17/0.44,3.04/0.56,2.91/0.48,2.78/0.37}
    \fill[LIFlowGray!65!white] (15.02,\y) rectangle ++(\w,0.035);
  \node[font=\fontsize{7.5pt}{9pt}\selectfont\sffamily]
    at (13.4,2.36) {Page image};
  \node[font=\fontsize{7.5pt}{9pt}\selectfont\sffamily]
    at (15.3,2.36) {Transcription};

  \draw[real flow] (check.east) -- (11.7,4.9) -- (11.7,3.65) -- (12.4,3.65);
  \draw[draw=DOSynthetic, line width=0.8pt, rounded corners=4pt]
    (page.east) -- (11.4,2.0) -- (11.4,0.65) -- (target.east);
  \fill[DOSynthetic] (11.4,1.325) circle (1.25pt);
  \draw[synthetic flow] (11.4,1.325) -- (11.7,1.325) -- (11.7,2.65) -- (12.4,2.65);
\end{tikzpicture}
\endgroup
\caption{\textbf{How the training set is built.} For real pages, we generate transcriptions, check them against the page image and available PDF text, and correct errors before accepting them as ground truth. For synthetic pages, we generate the content first and render it using styles from real documents, so the correct transcription is already known.}
\label{fig:data-overview}
\end{figure}

\subsection{Building the corpus}

\label{corpus}
We built the corpus targeting the areas where document-parsing VLMs still fail, including dense and spanned tables, mathematics, non-Latin scripts, old scans and handwriting, and charts.
We iteratively refine the SFT dataset, using model evaluations to guide targeted data collection and synthesis.
We retain data from earlier rounds to preserve existing capabilities.
The training set has 100,979 pages, about 0.37\,B target tokens: 82,943 real pages and 18,036 synthetic pages ($\approx$121\,M target tokens).
Real and synthetic pages share the ground-truth format shown in \Cref{appendix:annotations}.

\textbf{Real pages} come from eleven source families, chosen for breadth of layout and a permissible license (\Cref{tab:real-sources}).
We collected about one million PDFs from these sources and kept the pages counted in \Cref{tab:real-sources}.
To ensure diversity, we take at most three pages per document (six for table collections) and let no single source family, language or query group dominate.
We use the PDF text layer to filter for pages that contain tables, charts and mathematics.

\textbf{Synthetic pages} let us generate challenging examples at scale with exact ground truth.
Long tables, for example, are hard to find in real collections and difficult to annotate accurately.
Pages emphasizing each element type are produced by a dedicated generator (\Cref{tab:synth-sources}).
We ensure that the generators produce diverse formatting that still reflects what real documents look like.

\begin{table}[t]
\RCsetup
\begin{tabularx}{\textwidth}{>{\raggedright\arraybackslash}p{4.3cm}r>{\raggedright\arraybackslash}X}
\toprule
\RChead{Source} & \RChead{Pages} & \RChead{Document class} \\
\midrule\RCstart
Internet Archive & 23,722 & public-domain scans (1800-1975 books on science and mathematics, statistical annexes, encyclopaedias, and letters) and magazines \RCnext
arXiv & 16,142 & scientific formula-dense pages \RCnext
US / UK government statistical publications & 12,966 & Federal Reserve, IRS SOI, Census, FDIC, BEA, EPA, USDA, EIA. \RCnext
PubMed Central & 6,988 & flowcharts (PRISMA, decision trees), scientific articles \RCnext
Contracts and public documents with tracked changes & 6,903 & CUAD contracts, UN and EMA public documents \RCnext
Library of Congress & 5,199 & handwritten manuscripts and early printed matter with volunteer transcriptions \RCnext
OpenAlex-indexed articles & 2,447 & rare languages, tables in Arabic / Chinese / Korean / Japanese / Persian, charts \RCnext
EUR-Lex & 1,720 & 22 non-English language editions \RCnext
govinfo, NTRS, ERIC, DailyMed, SEC EDGAR, municipal ACFRs, BIS, EDINET & 5,879 & government and financial reports, born-digital tables, NASA and education scans \RCnext
World Bank Open Knowledge Repository & 1,479 & chart-dense reports \RCnext
Publisher sites & 492 & magazines \RClast
\bottomrule
\end{tabularx}
\caption{Real pages collected by source. We exclude 994 of these pages from training because they exceed the context limit.}
\label{tab:real-sources}
\end{table}

\begin{table}[t]
\RCsetup
\centering\begin{tabularx}{0.55\textwidth}{>{\raggedright\arraybackslash}Xr}
\toprule
\RChead{Element type} & \RChead{Pages} \\
\midrule\RCstart
Tables and charts & 13,953 \RCnext
Forms and formatting & 3,189 \RCnext
Text variants & 894 \RClast
\bottomrule
\end{tabularx}
\caption{We focus synthetic generation on tables and charts to obtain complex examples with exact ground truth.}
\label{tab:synth-sources}
\end{table}

\begin{figure}[t]
\centering

\newsavebox{\DXrowbox}
\newcommand{\DXrow}[1]{\sbox{\DXrowbox}{#1}%
  \ifdim\wd\DXrowbox>\linewidth\resizebox{\linewidth}{!}{\usebox{\DXrowbox}}\else\makebox[\linewidth]{\usebox{\DXrowbox}}\fi}
\newcommand{\DXgap}{\hspace{0.8em}}
{\raggedright\small\sffamily\bfseries Real pages\par}\vspace{2pt}
\DXrow{\begin{tabular}[t]{@{}c@{}}\fbox{\includegraphics[height=4.4cm]{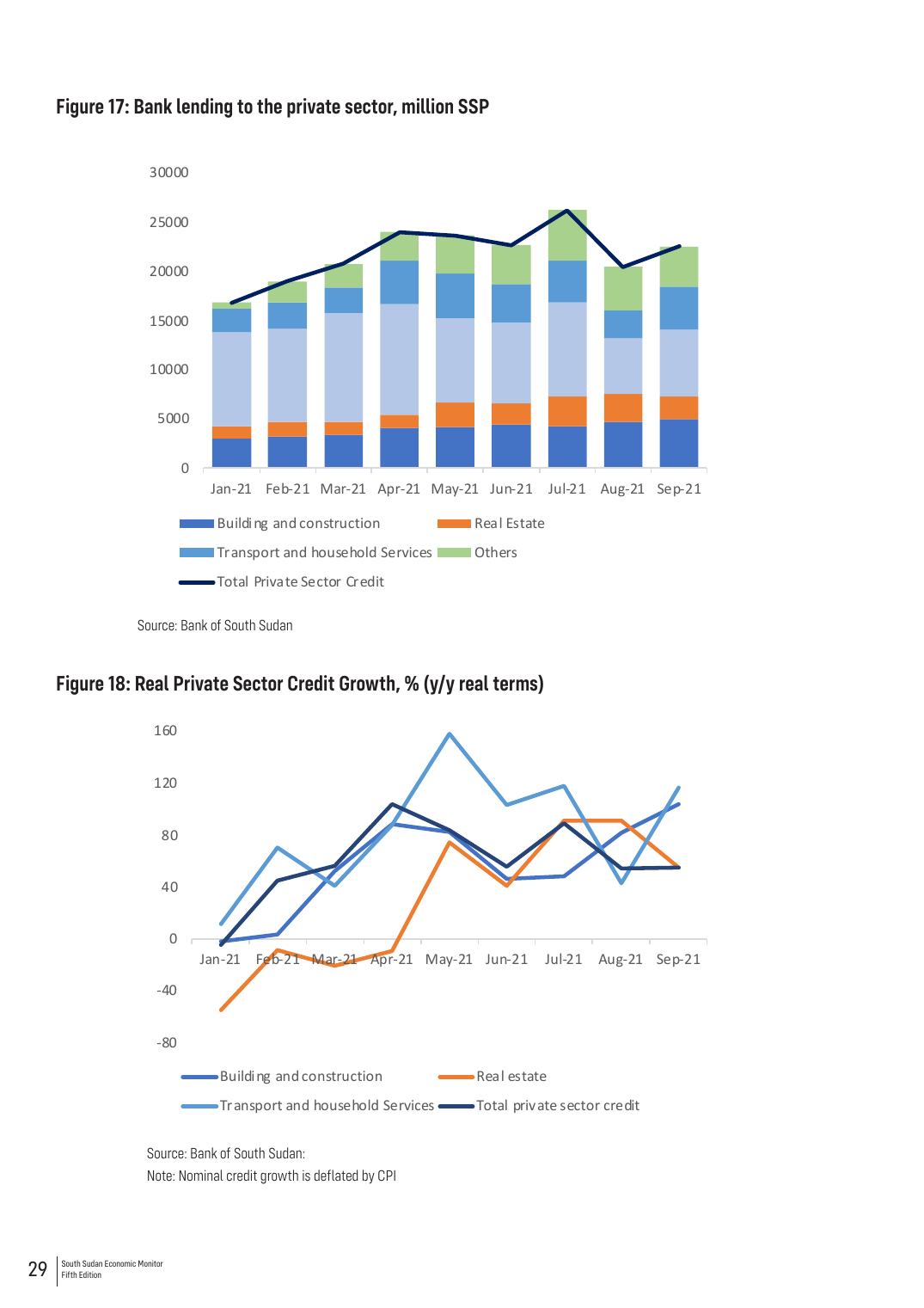}}\\[2pt]
  {\footnotesize\sffamily Chart-dense report page}\end{tabular}\DXgap\begin{tabular}[t]{@{}c@{}}\fbox{\includegraphics[height=4.4cm]{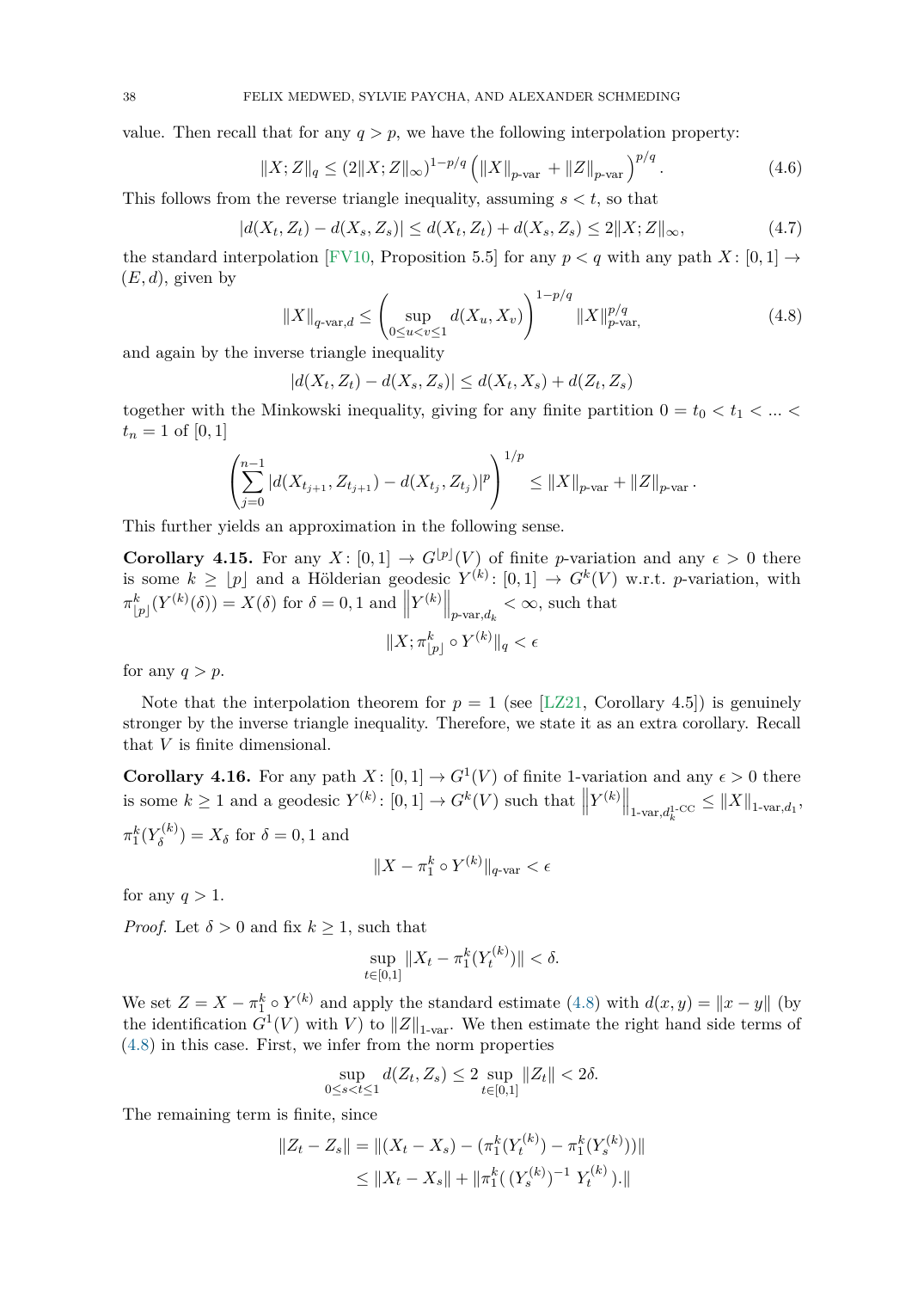}}\\[2pt]
  {\footnotesize\sffamily arXiv, formulas}\end{tabular}\DXgap\begin{tabular}[t]{@{}c@{}}\fbox{\includegraphics[height=4.4cm]{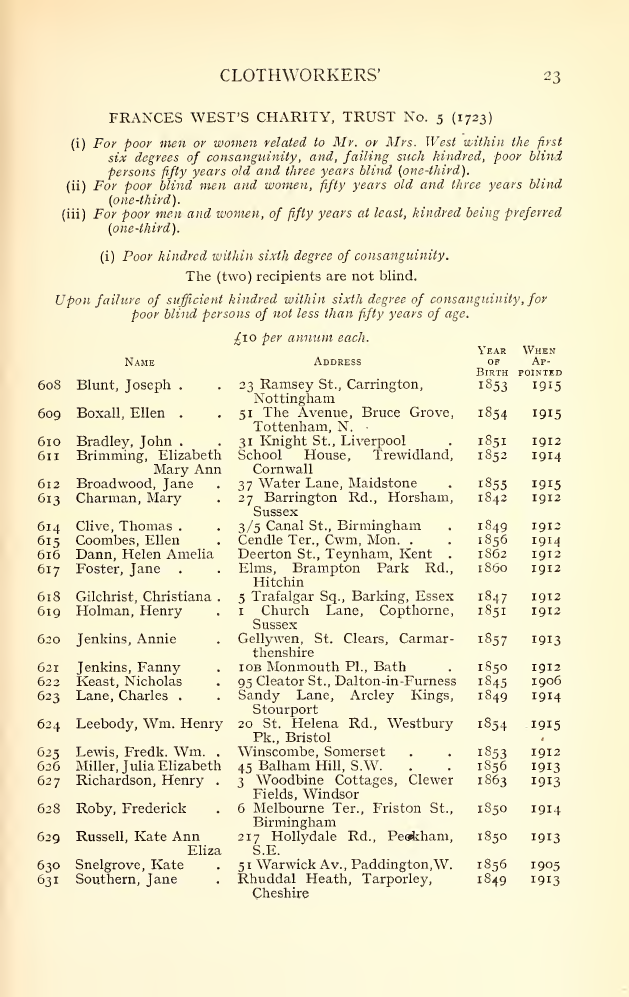}}\\[2pt]
  {\footnotesize\sffamily Internet Archive, scan}\end{tabular}\DXgap\begin{tabular}[t]{@{}c@{}}\fbox{\includegraphics[height=4.4cm]{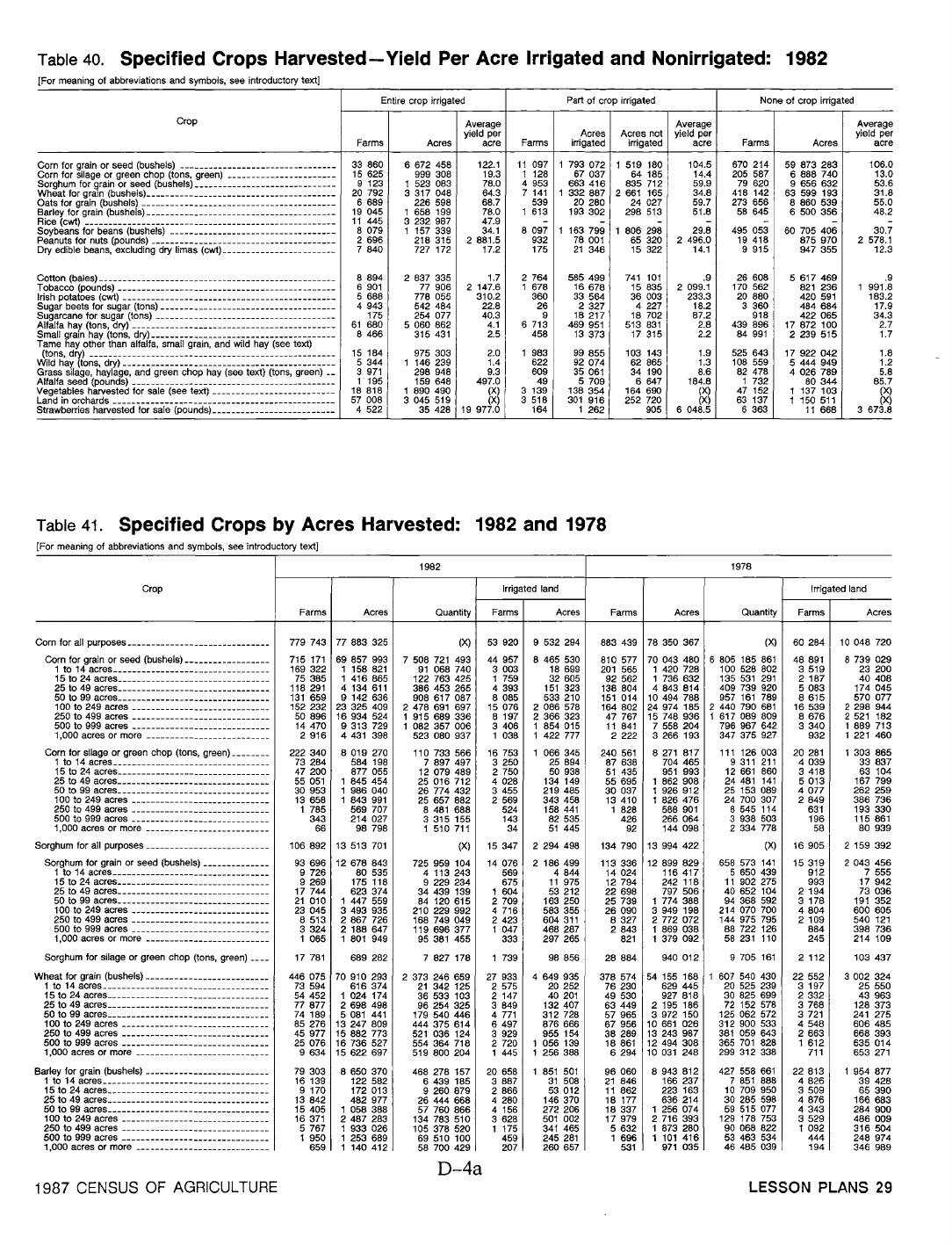}}\\[2pt]
  {\footnotesize\sffamily Government statistical table}\end{tabular}}\par\vspace{6pt}
{\raggedright\small\sffamily\bfseries Synthetic pages\par}\vspace{2pt}
\DXrow{\begin{tabular}[t]{@{}c@{}}\fbox{\includegraphics[height=4.4cm]{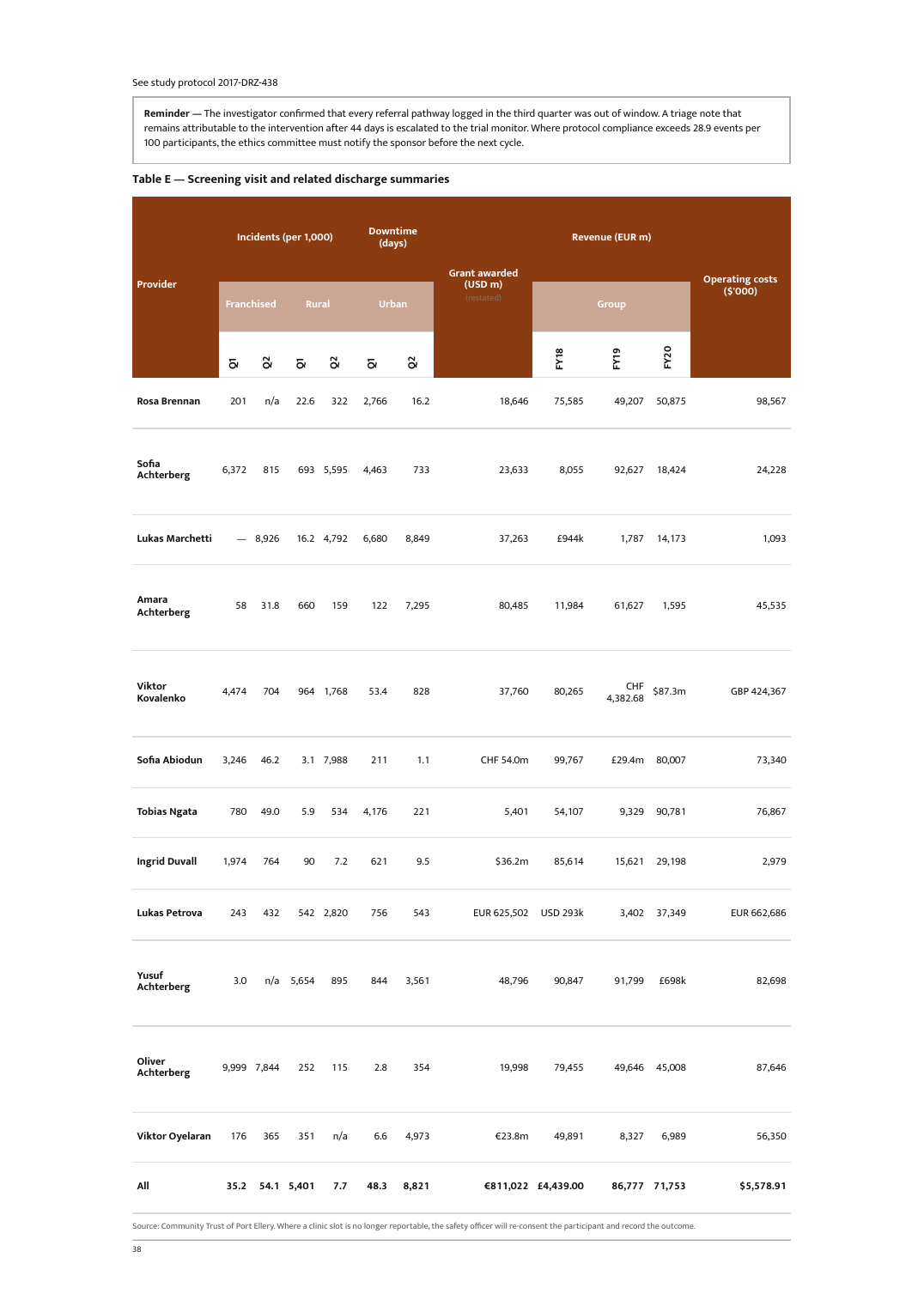}}\\[2pt]
  {\footnotesize\sffamily Spanned table}\end{tabular}\DXgap\begin{tabular}[t]{@{}c@{}}\fbox{\includegraphics[height=4.4cm]{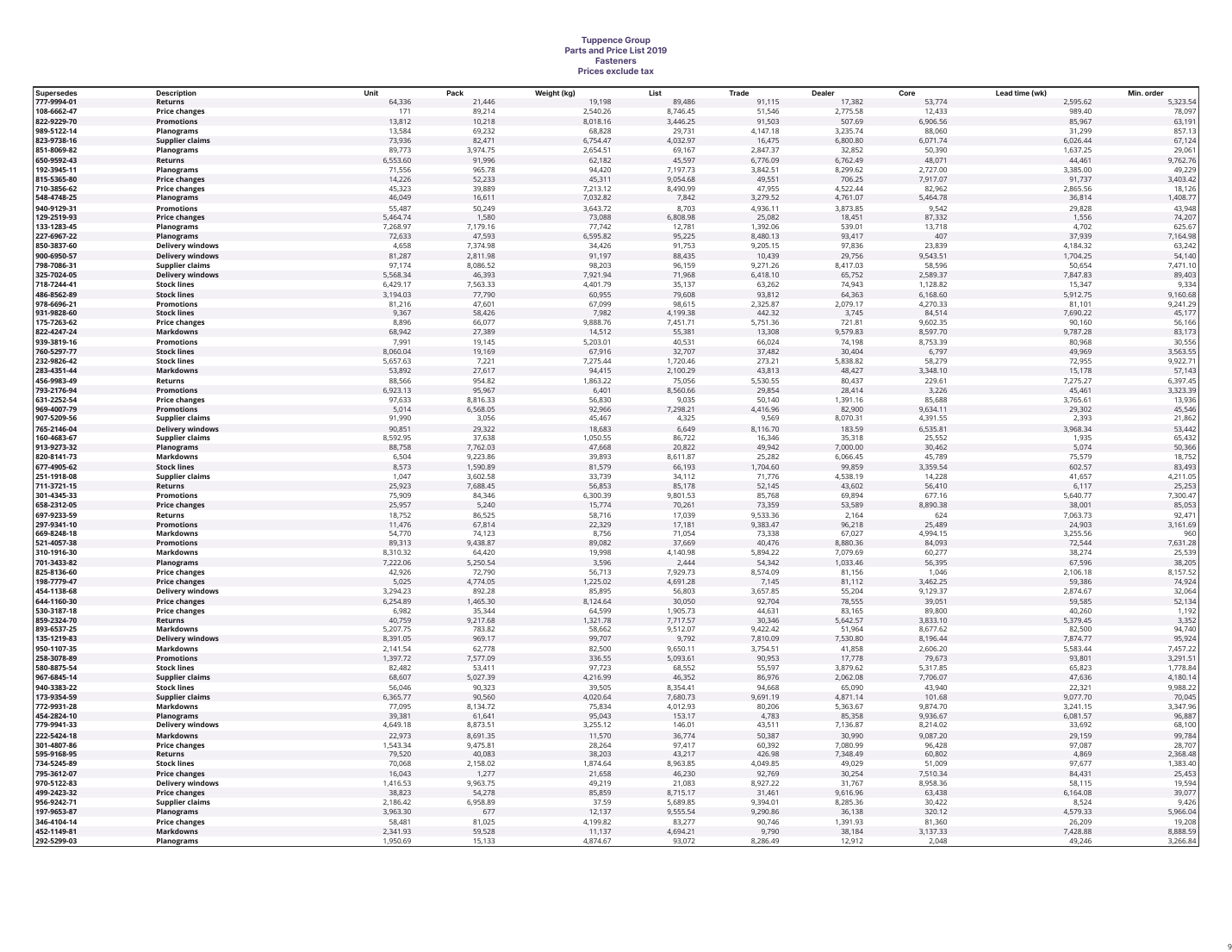}}\\[2pt]
  {\footnotesize\sffamily Very long, dense table}\end{tabular}\DXgap\begin{tabular}[t]{@{}c@{}}\fbox{\includegraphics[height=4.4cm]{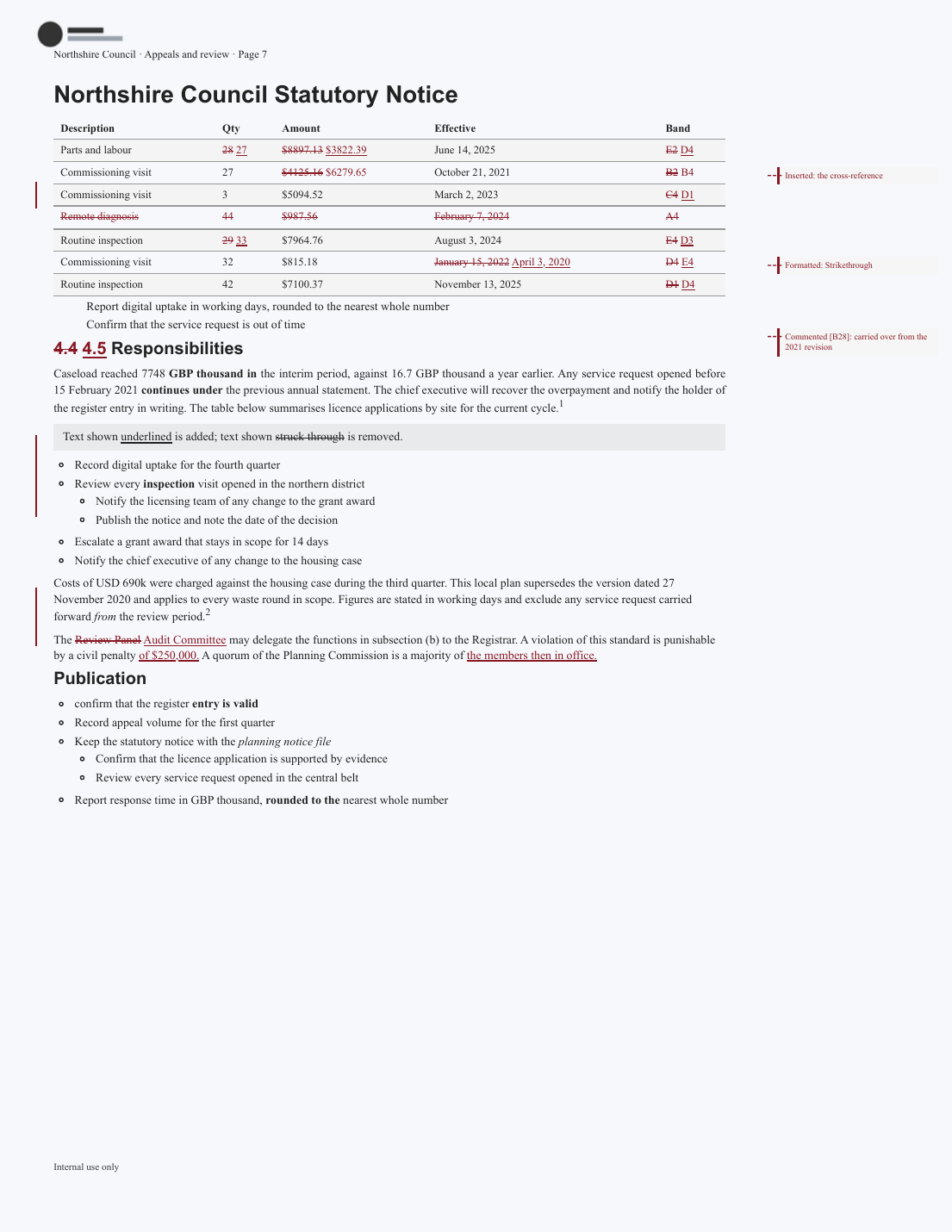}}\\[2pt]
  {\footnotesize\sffamily Text with tracked changes}\end{tabular}\DXgap\begin{tabular}[t]{@{}c@{}}\fbox{\includegraphics[height=4.4cm]{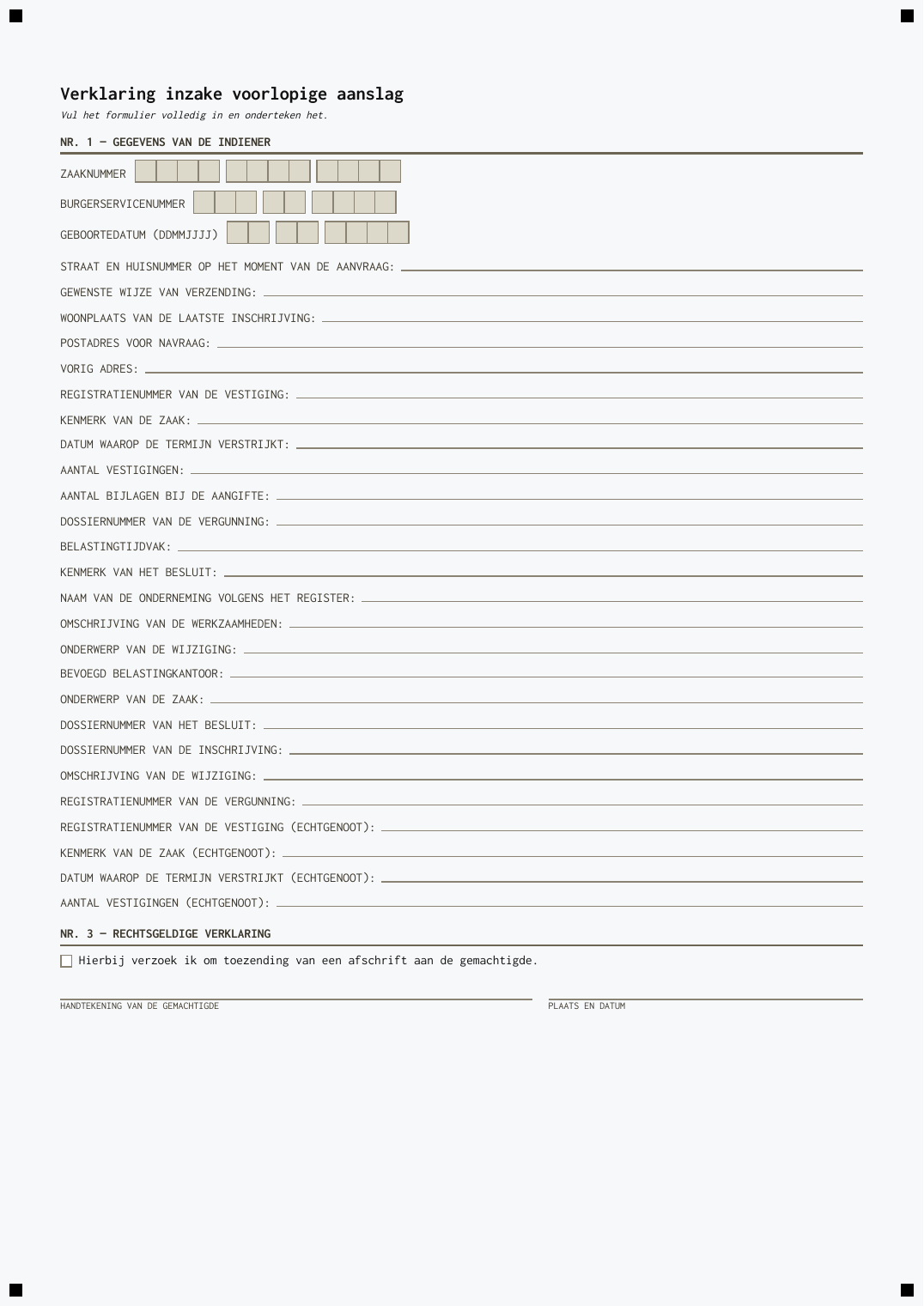}}\\[2pt]
  {\footnotesize\sffamily Form with fields}\end{tabular}}

\caption{\textbf{Example training pages.} Top: real pages from the chart, arXiv, Internet Archive and government statistical collections. Bottom: synthetic pages from the spanned-table, very-long-table, tracked-changes and form generators.}
\label{fig:data-examples}
\end{figure}

\subsection{Transcribing and verifying real pages}
\label{annotations}
\label{verification}

We use our proprietary ground truth generation workflow to produce accurate ground truth for SFT at a large scale.
The workflow consists of the following steps (\Cref{fig:data-overview}).

\paragraph{Teacher annotations.}

The teacher is composed of workflows to parse different complex page elements such as tables and charts, whose outputs are then combined with text predictions into the transcription of the entire page.
The transcription follows the output format described in \Cref{appendix:annotations}.

\paragraph{Annotation audit.}

Since teacher annotations might not be perfect, once the initial transcription of a page has been produced, a mid-size VLM checks that the transcription is correct with respect to the source page image using a set of auditing rules, which target different aspects of the transcription (\Cref{app:verification}).

The rules check for various common failure modes, such as incorrect table headers and table cells holding the wrong value.
When a rule fails, the VLM is called to correct the transcription, focusing on the detected issues.

\paragraph{Text layer grounding.}
For digital PDF pages in which the text content can be extracted directly, the ground truth is compared against the PDF text to ensure the completeness of the transcription.
While this text layer lacks the semantic structure of the page, the ground truth needs to at least match the text for it to be correct.

\paragraph{Correction \& re-evaluation.}

In correction \& re-evaluation, we use a large VLM to correct the page transcription and re-evaluate the correctness of the generated ground truth.

A digital PDF page is approved as ground truth when it passes both the audit and the text layer grounding; a page that fails either check, and every scanned page, goes to correction \& re-evaluation (\Cref{app:verification}).

Only when the large VLM's validation and correction attempts fail do we route the page for human revision.

\subsection{Generating synthetic pages}
\label{synthetic}

Real documents best represent what the model will see in use.
However, some of the complex document types that these models struggle with are scarce, and for others precise ground truth cannot be extracted reliably at scale.
To close those gaps, we use synthetic data that follows the distribution of real documents while keeping a high diversity of styles.

We developed an internal workflow that generates PDF pages from pre-synthesised ground truth: the generator samples document styles and templates from real documents while rendering that ground truth.
The generators cover tables and charts, including long tables of 5k--12k tokens, spanned headers, icon cells and 28 chart types; forms and inline formatting, including checkboxes and fill-in fields; and documents shown with tracked changes (\Cref{tab:synth-sources}).
The synthetic ground truth is generated in the same format as the ground truth of the real pages (\Cref{appendix:annotations}).

\section{Training}
\label{training}
\label{sft}
\label{rl}

We train \ocria{} in two stages for accurate document parsing.
SFT teaches Qwen3.5-4B to transcribe document pages in our output format.
RLVR addresses errors left by SFT by scoring page structure and content directly, such as table cells, chart values and formatting marks, while penalizing degenerate output such as repetition.
A separate draft model accelerates inference (\Cref{inference}).

\paragraph{SFT.}

We fine-tune Qwen3.5-4B on 100,979 pages (\Cref{data}).
We prompt the model to transcribe each page (\Cref{app:prompts}), using Markdown with HTML tables to represent its content and structure.

We render each page at 1,024\,px on its long side and use a 20,480-token context to accommodate long tables.
\Cref{app:sft-setup} gives the filtering and training settings; \Cref{app:train-resolution} reports results at higher image resolutions.

\paragraph{RLVR.}

After SFT, we train the model with GRPO \citep{shao2024deepseekmath,guo2025deepseekr1} with the goal of refining and enhancing the behaviors
that the model has already learned to some extent.
We also want it to improve its formatting and consistency.
Our RLVR setup differs from previous work in two ways. First, it uses a larger set of rewards covering different page elements, along with additional penalties. Second, it weights the rewards dynamically according to their variance across rollouts, so that page elements that appear less often still contribute meaningfully to the update.
\Cref{app:rl-setup} gives the settings.

\begin{itemize}
\item \textbf{Training data.}
The stage trains on 11,234 pages (\Cref{tab:grpo-data}).
For rollout generation we keep the image at its stored resolution, mostly at or above 150\,dpi (\Cref{tab:train-image-dpi}).
We train on the full corpus without removing pages with low GRPO advantages, as removing such pages gave mixed results (\Cref{app:page-selection}).

\item \textbf{Reward design.}
Our reward function combines terms across multiple dimensions of transcription quality (\Cref{app:reward-comparison}), giving partial credit for correct aspects of imperfect outputs.
Most of its thirty-four terms (\Cref{app:reward-components}) are scores, which give credit for what is right.
Twelve are penalties, which are zero unless the output has the error they check for, e.g., a penalty for merged or split tables (\Cref{app:reward-formulas}).

To ensure the reward function is well defined and hard to game, we test each reward term against ground truth before training (\Cref{app:reward-tests}).

\item \textbf{Reward weighting.}
We carefully weight reward terms with variance-based reweighting to balance their influence on group-relative updates.
A term's influence depends on both its weight and its variation across rollouts, so fixed weights alone can let one
dimension dominate at the expense of others (\Cref{app:reward-combining}).
We also avoid normalizing each term by its own within-group statistics \citep{mogrpo2025}, which can give penalties as
much influence as the quality terms they are intended to protect.

\end{itemize}

We train several checkpoints that are stronger in different dimensions and merge them into the final \ocria{} model by checkpoint averaging \citep{wortsman2022soups,rame2023rewardedsoups}.

\section{Inference}
\label{inference}
\label{speculator}

We accelerate inference through speculative decoding, mixed-precision quantization, and optimizations to the vLLM serving runtime.

\paragraph{Speculative decoding.}
We use speculative decoding~\citep{leviathan2023fast,chen2023accelerating} for a lossless inference speedup: a small draft model proposes several tokens ahead and the full model verifies them all in one forward pass.
We trained a six-layer DFlash~2~\citep{dflash2026,inco2026dflash2} block-diffusion speculator with 0.69B parameters on document pages from our training corpus.
We use a block of 32 tokens, against the 8 to 16 tokens used by DFlash and by the draft of HunyuanOCR-1.5~\citep{hunyuanocr2026}, so a verification step can produce up to 32 tokens.

We implemented a dynamic draft length, using longer drafts at low concurrency to improve per-request decoding speed and shorter drafts at higher concurrency to preserve throughput as verification costs grow.
With 31 draft tokens, each verification step emits an average of 12.5 to 17.5 tokens depending on the page set, including one token from the target.
These averages remain consistent across evaluated batch sizes (\Cref{tab:acceptance}).

We also refined DFlash~2's draft-token selection with dynamic programming, slightly increasing accepted tokens per step at negligible cost.
\Cref{app:speculator} provides further implementation details.

\paragraph{Quantization.}
We quantize \ocria{} to a mix of NVFP4 and FP8~\citep{nvidia2025nvfp4}, while retaining BF16 for the vision encoder.
We calibrate on the model's own transcriptions and select each layer's precision, keeping only configurations whose scores remain within measurement noise of the BF16 baseline.

\paragraph{Serving optimizations.}
We customize vLLM~\citep{kwon2023vllm} to reduce repeated work and improve GPU utilization.
Our port of ReplaySSM~\citep{vllm2026replayssm} replays accepted tokens from a saved recurrent state, avoiding a state snapshot at every draft position.
We write several CUDA kernels to optimize inference.
We also fuse activation and quantization operations, build batch metadata once per step, and reuse the shared prompt prefix across requests.
\Cref{app:serving} gives the quantization, implementation and validation details.

\section{Evaluation}
\label{evaluations}

We evaluate \ocria{} as a document-parsing VLM along two axes: parsing quality and serving speed.
We measure parsing quality across four public benchmarks with complementary coverage, from enterprise pages and broad document layouts to cropped tables in nine languages; we measure serving speed on one B200 GPU.
We first describe the benchmark scope and the evaluation protocol (\Cref{eval:protocol}), then report parsing results (\Cref{eval:main}), and finally measure inference speed (\Cref{eval:speed}).

\subsection{Benchmarks and protocol}
\label{eval:protocol}

We have considered four well-known public benchmarks.
The benchmarks rely on different document distributions due to their focus, and this variety ensures a thorough evaluation of our model across a diversity of document parsing problems.
Unless otherwise specified, and to ensure fair evaluation, we have evaluated \ocria{} with the transcription prompt of \Cref{app:prompts}, on the same page renderings at each benchmark's standard resolution, and scored it with the benchmark's own scorer. \Cref{app:eval-protocol} provides additional details about the evaluation protocol.
The benchmarks are:

\begin{itemize}

\item ParseBench~\citep{zhang2026parsebenchdocumentparsingbenchmark} contains pages from enterprise documents such as insurance, finance and government filings, and grades text content, text formatting and charts with rules written against each page, and tables with a table-structure metric (\Cref{app:eval-protocol}).
Its fifth dimension, visual grounding, scores whether each extracted block is located, labelled and matched to its content; \ocria{} does not output bounding boxes at this moment, so we leave that dimension out.

\item olmOCR-bench~\citep{poznanski2025olmocrunlockingtrillionstokens} runs 7,010 pass-or-fail unit tests on 1,403 pages, grouped into arXiv mathematics, old scans with mathematics, old scans, tables, multi-column pages, headers and footers, and long passages of small text.

\item PulseBench-Tab~\citep{pulsebenchtab2026} scores 1,820 cropped tables in nine languages by comparing the predicted cell grid with the reference grid.

\item fr-bench-pdf2md~\citep{frbenchpdf2md2026} runs olmOCR-style tests on 721 difficult French pages.

\end{itemize}

\subsection{Main results}
\label{eval:main}

\ocria{} has the highest score on every benchmark in \Cref{tab:main-results}, and its lead comes from the content we focused on: tables, charts and text formatting.
Its margin is largest on charts and text formatting, where some other systems score close to zero, and on plain text it is as accurate as the best system on ParseBench.

\begin{table}[!htbp]
\RCsetup
\begin{tabularx}{\linewidth}{Xcccc}
\toprule
\RChead{Model} & \RChead{ParseBench$^{*}$} & \RChead{olmOCR-bench$^{\dagger}$} & \RChead{PulseBench-Tab} & \RChead{fr-bench-pdf2md} \\
\midrule\RCstart
Infinity-Parser2-Pro (35B) & 0.74 & 90.2 & 0.82 & 54.9 \RCnext
LightOnOCR-3-4B & 0.75 & 90.1 & 0.82 & 71.4 \RCnext
olmOCR-2 (7B) & 0.43 & 83.1 & 0.80 & 56.9 \RCnext
Chandra-OCR-2 (4B) & 0.75 & 89.4 & 0.80 & 65.5 \RCnext
dots.mocr (3B) & 0.56 & 87.4 & 0.82 & 56.8 \RCnext
Surya OCR 2 (0.7B) & 0.58 & 85.4 & 0.71 & 50.9 \RCnext
LightOnOCR-2-1B & 0.61 & 88.1 & 0.81 & 53.5 \RCnext
Mistral OCR 4.1 & 0.68 & 80.3 & 0.75 & 54.1 \RClast
\midrule\RCstart
\ocria{} & \textbf{0.87} & \textbf{90.5} & \textbf{0.84} & \textbf{72.1} \RClast
\bottomrule
\end{tabularx}

{\par\smallskip\raggedright\footnotesize
$^{*}$Mean over text content, text formatting, tables and charts; excludes visual grounding.\par
$^{\dagger}$For olmOCR-bench, the mean of its three categories on tables and formulas (Tables, arXiv math and Old scans math); \Cref{tab:olmocr-categories} gives the official score over all eight categories. \Cref{app:eval-breakdown} breaks down olmOCR-bench, PulseBench-Tab and fr-bench-pdf2md by category.\par}
\caption{\ocria{} leads on four well-known benchmarks, especially in tables, charts and text formatting.}
\label{tab:main-results}
\end{table}

\begin{table}[!htbp]
\RCsetup
\begin{tabularx}{\linewidth}{Xccccc}
\toprule
\RChead{Model} & \RChead{Mean$^{*}$} & \RChead{Tables} & \RChead{Charts} & \RChead{Text formatting} & \RChead{Text content} \\
\midrule\RCstart
Infinity-Parser2-Pro (35B) & 0.74 & 0.86 & 0.61 & 0.59 & \textbf{0.90} \RCnext
LightOnOCR-3-4B & 0.75 & 0.83 & 0.66 & 0.64 & 0.89 \RCnext
olmOCR-2 (7B) & 0.43 & 0.73 & 0.14 & 0.02 & 0.84 \RCnext
Chandra-OCR-2 (4B) & 0.75 & 0.89 & 0.65 & 0.61 & 0.84 \RCnext
dots.mocr (3B) & 0.56 & 0.85 & 0.00 & 0.47 & \textbf{0.90} \RCnext
Surya OCR 2 (0.7B) & 0.58 & 0.80 & 0.01 & 0.68 & 0.84 \RCnext
LightOnOCR-2-1B & 0.61 & 0.78 & 0.15 & 0.64 & 0.88 \RCnext
Mistral OCR 4.1 & 0.68 & 0.74 & 0.40 & 0.66 & \textbf{0.90} \RClast
\midrule\RCstart
\ocria{} & \textbf{0.87} & \textbf{0.92} & \textbf{0.85} & \textbf{0.83} & \textbf{0.90} \RClast
\bottomrule
\end{tabularx}
{\par\smallskip\raggedright\footnotesize
$^{*}$Mean over the four dimensions shown; excludes visual grounding.\par}
\caption{On ParseBench, \ocria{} leads on tables, charts and text formatting, and ties for best on text content.}
\label{tab:parsebench-dims}
\end{table}

On ParseBench it scores 0.85 on charts against 0.66 for the next-best system, 0.83 on text formatting against 0.68, 0.92 on tables against 0.89, and 0.90 on text content, tied with the best other systems (\Cref{tab:parsebench-dims}).
It also scores highest on the two other benchmarks that contain these elements: PulseBench-Tab, which contains only tables (0.84 against 0.82), and fr-bench-pdf2md, which contains forms, graphics and long tables (72.1 against 71.4, \Cref{tab:main-results}).

olmOCR-bench has no test of charts or text formatting, so for it \Cref{tab:main-results} gives the mean of the three categories that test tables and formulas, two of the elements \ocria{} is built to transcribe.
\ocria{} has the highest score on these categories, 0.3 points ahead of Infinity-Parser2-Pro (\Cref{tab:olmocr-categories}).
In the additional comparison in \Cref{tab:olmocr-same-render}, \ocria{} also has the highest score over all eight categories, again by 0.3 points.
This comparison combines our runs at 1,288\,px with published scores for Mistral OCR 4.1~\citep{mistral2026ocr41} and olmOCR-2~\citep{poznanski2025olmocr2unittest}.

On PulseBench-Tab \ocria{} also leads on the other three measures of \Cref{tab:pulse}: its median table score is 0.95 against 0.93 for the next-best system, it gives a parseable table for 99.1\% of the tables against 97.7\%, and it reproduces 485 of the 1,820 tables perfectly against 431.

\Cref{app:examples} shows outputs of \ocria{} on tables, charts, text formatting and handwritten formulas.
Also, \Cref{eval:failures} shows cases where \ocria{}'s performance is lower. Most of the low-performing cases are the hardest cases within the areas we targeted, e.g. old scanned documents and less frequent languages.

\subsection{Inference efficiency}
\label{eval:speed}

For inference speed, we use a separate comparison set of fast open-weight document-parsing VLMs, including compact models and systems with speculative decoding.
On one B200, \ocria{} leads the five evaluated systems in both decoding speed and page throughput at one and eight concurrent requests (\Cref{fig:speed-systems}).
Its per-request decoding speed is respectively $3.4\times$ and $2.0\times$ that of the next-fastest system.

In a separate concurrency sweep, \ocria{} parses 10.21 pages per second at 64 concurrent requests, the highest concurrency tested with p90 latency below 10\,s (\Cref{tab:speed-results}).
We evaluate the serving stack described in \Cref{inference}; \Cref{app:speed-protocol} gives the measurement protocol, full results and comparison with the BF16 checkpoint on stock vLLM.

\begin{figure}[!htbp]
\centering
\input{figures/speed_systems_style}
\begin{tikzpicture}
\begin{axis}[speedsys, speedsyslabels, name=axA, title={(a) One request: pages/s}, xlabel={pages/s},
  xmax=2.6, nodes near coords={\pgfmathprintnumber[fixed, precision=2, fixed zerofill]{\pgfplotspointmeta}}]
\addplot[speedsysocria] coordinates {(2.01,ocria)};
\addplot[speedsysother] coordinates {(0.80,nemotron) (0.71,hunyuan) (0.55,smol) (0.39,falcon)};
\end{axis}
\begin{axis}[speedsys, yticklabels={}, name=axB, at={(axA.outer north east)}, anchor=outer north west, xshift=1mm,
  title={(b) One request: decoding speed}, xlabel={output tok/s per request},
  xmax=5800, xtick={0,2000,4000}, xticklabel style={/pgf/number format/.cd, fixed, 1000 sep={{,}}},
  nodes near coords={\pgfmathprintnumber[fixed, precision=0, 1000 sep={{,}}]{\pgfplotspointmeta}}]
\addplot[speedsysocria] coordinates {(4734,ocria)};
\addplot[speedsysother] coordinates {(1395,nemotron) (1403,hunyuan) (894,smol) (599,falcon)};
\end{axis}
\end{tikzpicture}\par
\vspace{0.7em}
\input{figures/speed_systems_style}
\begin{tikzpicture}
\begin{axis}[speedsys, speedsyslabels, name=axA, title={(c) Eight requests: pages/s}, xlabel={pages/s},
  xmax=9, nodes near coords={\pgfmathprintnumber[fixed, precision=2, fixed zerofill]{\pgfplotspointmeta}}]
\addplot[speedsysocria] coordinates {(7.83,ocria)};
\addplot[speedsysother] coordinates {(4.75,nemotron) (3.26,hunyuan) (3.21,smol) (1.95,falcon)};
\end{axis}
\begin{axis}[speedsys, yticklabels={}, name=axB, at={(axA.outer north east)}, anchor=outer north west, xshift=1mm,
  title={(d) Eight requests: decoding speed}, xlabel={output tok/s per request},
  xmax=2400, xtick={0,1000,2000}, xticklabel style={/pgf/number format/.cd, fixed, 1000 sep={{,}}},
  nodes near coords={\pgfmathprintnumber[fixed, precision=0, 1000 sep={{,}}]{\pgfplotspointmeta}}]
\addplot[speedsysocria] coordinates {(1932,ocria)};
\addplot[speedsysother] coordinates {(963,nemotron) (787,hunyuan) (732,smol) (356,falcon)};
\end{axis}
\end{tikzpicture}
\caption{\ocria{} leads in page throughput (left) and decoding speed per request (right) at one (top) and eight (bottom) concurrent requests. Each system runs on one B200. \Cref{tab:speed-systems} gives the exact values.}
\label{fig:speed-systems}
\end{figure}

\FloatBarrier

\section{Related Work}
\label{related}

\ocria{} reads the whole page with one VLM, and it differs from earlier document-parsing VLMs mainly in its training data and its rewards.
The document-parsing VLMs we compare against include olmOCR-2~\citep{poznanski2025olmocr2unittest}, dots.mocr~\citep{zheng2026dotsmocr}, LightOnOCR~\citep{taghadouini2026lightonocr1bendtoendmultilingual,lightonocr3blog2026}, Chandra~\citep{datalab2026chandra}, Infinity-Parser2-Pro~\citep{huang2026infinityparser2technicalreport} and Surya~\citep{paruchuri2025surya}, and other recent ones include the DeepSeek-OCR series~\citep{wei2025deepseekocrcontextsopticalcompression,wei2026deepseekocr2visualcausal} and TeleOCR~\citep{cai2026teleocrnavigatingdocumentparsing}.
Pages with correct annotations are scarce, so many training sets are labeled by a larger model or generated synthetically.
These models are weakest on the page types with the fewest annotated examples, such as charts, long or heavily spanned tables, low-resource languages and scanned pages.
We built our training data around these page types, and we believe this data is one of the reasons \ocria{} outperforms the models we compare it with (\Cref{data,evaluations}).

Some of these models continue training with reinforcement learning on a reward computed from the parsed output~\citep{wang2025infinityparserlayoutaware,poznanski2025olmocr2unittest,taghadouini2026lightonocr1bendtoendmultilingual, huang2026infinityparser2technicalreport}.
Compared to previous work, our RLVR has two main differences.
The first is a larger set of rewards for different page elements and more penalties (\Cref{app:reward-comparison}).
The second is a mechanism to dynamically weight the different rewards based on rollout variance (\Cref{app:reward-combining}), which gives a higher contribution to less frequent and varied rollout elements.

We combine a block-diffusion draft model with quantization and custom kernels, and we report the resulting inference performance (\Cref{eval:speed}).
There are other models using several techniques to speed up inference, such as
GLM-OCR~\citep{duan2026glmocrtechnicalreport}, which predicts several tokens per step, and diffusion-based models~\citep{man2026dododiscreteocrdiffusion,dong2026minerudiffusion}, which write many tokens in parallel.
Some document-parsing VLMs also use speculative decoding: HunyuanOCR-1.5 with a DFlash draft model~\citep{hunyuanocr2026}, HPD-Parsing with multi-token prediction heads~\citep{hpdparsing2026}, and GravityOCR with drafts from the same network~\citep{gravityocr2026}; \Cref{app:speculator} compares how many draft tokens they accept per step with ours.

\section{Conclusion and Future Work}
\label{conclusion}

At the time of this report, \ocria{}, a 4B document-parsing VLM, has the highest scores on tables, charts and text formatting among all the systems we compare it with (\Cref{tab:main-results}), and its largest leads are on charts and text formatting in ParseBench (\Cref{tab:parsebench-dims}).
At ten pages per second, one GPU parses about 860,000 pages a day (\Cref{eval:speed}), so the training data for the next model, the backlog of a document store or the attachments a document agent receives can be parsed on one machine.
We plan to make a version of \ocria{} available soon.

Most of the work for \ocria{} went into preparing the training data.
Our focus on high quality data rather than quantity is one of the reasons of \ocria{}'s performance.
Synthetic pages came with ground truth from their generators, and real pages were transcribed by a multi-stage pipeline whose output we checked by rule and against the text stored in the PDF.
After SFT, we trained the model with RLVR and averaged checkpoints, and a draft model significantly speeds up decoding.

The same way of collecting and annotating data can be applied to prepare data for the cases where \ocria{} currently scores lowest, such as old scans, tables in Arabic, Greek and Korean, and long tables whose rows run past the output budget (\Cref{eval:failures}).
Also, we have not yet released a model capable of predicting layout, which is left as work for a future release.

\bibliographystyle{plainnat}
\bibliography{references}

\clearpage
\appendix
\section*{Appendix}
\crefalias{section}{appendix}
\crefalias{subsection}{appendix}

\section{Data}
\label{app:annotation-format}

This appendix gives the output format (\Cref{appendix:annotations}) and the verification rules (\Cref{app:verification}) behind \Cref{data}.

\subsection{Output format}
\label{appendix:annotations}
\begin{LIPanel}{Output format}
\renewcommand{\arraystretch}{1.15}
\begin{tabularx}{\linewidth}{@{}>{\raggedright\arraybackslash}p{2.9cm}>{\raggedright\arraybackslash}X@{}}
\LIRowLabel{Mathematics} &
  Inline formulas use \LIcode{$...$} and display formulas use \LIcode{$$...$$}; a printed equation number such as (3) is written as plain text after the closing \LIcode{$$}, not inside the formula. \\
\LIRowRule
\LIRowLabel{Tables \& Charts} &
  Write each table as one line of HTML with \LIcode{<th>} header cells and exact \LIcode{rowspan}/\LIcode{colspan} spans.
  Cells can contain \LIcode{<strong>}, \LIcode{<em>}, \LIcode{<u>}, \LIcode{<br/>} and \LIcode{<sup>}/\LIcode{<sub>}.
  Captions appear on the line before the table; charts are transcribed as tables. \\
\LIRowRule
\LIRowLabel{Inline formatting} &
  Use \LIcode{<sup>}/\LIcode{<sub>} tags rather than Unicode superscripts;
  \LIcode{**bold**}, \LIcode{*italic*}, \LIcode{<u>} and \LIcode{~~strike~~} mark bold, italic, underlined and struck-through text;
  \LIcode{<hr/>} marks a printed horizontal rule. \\
\LIRowRule
\LIRowLabel{Lists and forms} &
  List items use \LIcode{*} at the top level and \LIcode{-} for nested items; checkboxes use \LIcode{[x]}/\LIcode{[ ]}, and fill-in fields use \LIcode{[___]}. \\
\LIRowRule
\LIRowLabel{Page furniture} &
  Place the running header on the first line in \LIcode{<page_header>...</page_header>},
  the running footer on the last line in \LIcode{<page_footer>...</page_footer>},
  and the printed page number in \LIcode{<page_number>...</page_number>} inside its header or footer. \\
\LIRowRule
\LIRowLabel{Images} &
  Write \LIcode{![label: content]} without a filename: the label identifies the image type (logo, photo, chart, flowchart, \dots),
  and the content is a sentence describing a photograph, the printed words of a logo or seal, or the digits of a barcode.
  Signatures and seals use \LIcode{[signature: Name]} and \LIcode{[seal: text]};
  an overlay watermark appears as one \LIcode{<watermark>} line at the top. \\
\LIRowRule
\LIRowLabel{Flowcharts} &
  Place a fenced \LIcode{mermaid} block after the image marker. \\
\end{tabularx}
\end{LIPanel}

\subsection{Verification rules}
\label{app:verification}

For a digital PDF page, the transcription is approved as ground truth if it passes both the audit and the text layer grounding; if either fails, the page is sent to correction and re-evaluation.
A scanned page always goes to correction and re-evaluation, since there is no text layer to confirm the transcription.

\Cref{tab:verification-rules} lists the rules of the audit in \Cref{verification}, by group.
The audit and the reviewer answer every rule with pass, fail, uncertain or not applicable, and quote the part of the page that decided the answer.

\begin{table}[!htpb]
\RCsetup
\begin{tabularx}{\textwidth}{>{\raggedright\arraybackslash}p{2.7cm}>{\raggedright\arraybackslash}p{2.8cm}>{\raggedright\arraybackslash}X}
\toprule
\RChead{Group} & \RChead{Rules} & \RChead{What it catches} \\
\midrule
\RCgroup{Table rules}{25 rules, asked of every table}
Cells & T01--T04 & one wrapped cell split into several rows; separately printed records merged into one multi-line cell; invented spacer rows; real blank cells removed or filled \RCnext
Spans & T05, T07--T09, T23 & a category or subtotal label spanned across cells that hold values; wrong grid or span dimensions; repeated values merged into one cell; a real merged cell that loses what it covers; empty regions the rules treat as equivalent flagged as errors \RCnext
Headers & T10--T12 & header tiers flattened, dropped or made up; a header span over the wrong columns; missing or invented headers and columns \RCnext
Content & T06, T13--T16 & sparse or subsection values left out; a cell's text or number corrupted; superscripts, footnote marks or symbols detached from their value; a value in the wrong row or column; a long table that stops early or repeats \RCnext
Boundaries & T17--T22 & a borderless grid flattened into prose; one table split at a visual band; neighbouring tables woven together; a continuation that invents row relationships; printed occurrences duplicated or suppressed; captions, notes and totals attached to the wrong table \RCnext
Source image & T24, T25 & a source image or render that is incomplete or misleading; emphasis in the source turned into table structure \RClast
\RCgroup{Page rules}{7 rules, asked once per page}
Omissions & P01 & text outside the tables that is missing: labels, captions, notes, headers, footers \RCnext
Inventions & P02 & invented or duplicated text, including a plausible footer number \RCnext
Pictures & P03 & a picture that is unaccounted for or given the wrong kind \RCnext
Reading order & P04 & a wrong order across columns, sidebars or right-to-left text; tables and prose out of place \RCnext
Ownership & P05 & a caption, note, total or detached cell held by the wrong block at a table boundary \RCnext
Character fidelity & P06 & a character outside the tables that differs from the print, including a printed typo that was corrected \RCnext
Image completeness & P07 & a clipped or unreadable source image, or a missing illustration \RClast
\RCgroup{Script rules}{added for non-Latin pages}
Image markers & P08, P09 & a picture with no marker, two markers or the wrong class; a marker, header band or footer band that is not in the output format \RCnext
Script fidelity & P10 & an invented, dropped or substituted character; transliteration; digits converted to another numeral system (for example Arabic-Indic digits written as 0--9); a line repeated more often than it is printed \RClast
\bottomrule
\end{tabularx}
\caption{Rules of the audit. The table rules apply to every table of a draft and the page rules once to the page. The last group is added for pages in non-Latin scripts.}
\label{tab:verification-rules}
\end{table}

\section{Training}
\label{app:training}

This appendix gives the training details left out of the main text.
\Cref{app:sft-setup,app:rl-setup} list the SFT and RLVR settings, \Cref{app:train-resolution} reports the experiments with the training resolution, \Cref{app:reward-components} lists the components of the reward, and \Cref{app:page-selection} reports the tests on pages that give no training signal.

\subsection{Supervised fine-tuning settings}
\label{app:sft-setup}

The SFT context length is 20,480 tokens.
Context-length filtering excludes 994 real pages from the collected corpus, leaving 100,979 pages.

We keep the training recipe fixed when comparing data mixes to isolate their effects.
The SFT checkpoint for \ocria{} was trained for two epochs with AdamW at a global batch of 64 sequences, a peak learning rate of $1 \times 10^{-5}$, a cosine schedule with linear warm-up over 3\% of the steps (94 of 3,154), and zero weight decay.
Sequences use bf16 without sequence packing (each GPU processes two sequences at a time).
We chose the learning rate by starting from a setting that had already worked well (batch 16, learning rate $5 \times 10^{-6}$), scaling it with the square root of the batch size up to batch 64, and then keeping it fixed.

We train on four nodes with eight B200 GPUs each.

\subsection{Reinforcement-learning settings}
\label{app:rl-setup}

We train with verl~\citep{sheng2024hybridflow} and the settings in \Cref{tab:rl-setup}.
The trust region is asymmetric so that low-probability tokens have more room to rise \citep{yu2025dapo}.
The limits on the prompt and the sampled transcription add up to more than the 20,480-token context of the SFT stage, because a sampled transcription can be longer than its reference.

\begin{table}[t]
\RCsetup
\begin{tabularx}{\textwidth}{>{\raggedright\arraybackslash}p{5.2cm}X}
\toprule
\RChead{Setting} & \RChead{Value} \\
\midrule\RCstart
Rollouts per page & 12 \RCnext
Sampling & temperature 0.8, top-$p$ 0.9 \RCnext
Batch & 64 pages, 12 rollouts each (768 sequences) \RCnext
Optimiser & AdamW, learning rate $1\times10^{-6}$ \RCnext
Gradient clipping & 0.5 \RCnext
KL penalty & in the loss, coefficient 0.04 \RCnext
Trust region & 0.2 below, 0.28 above \RCnext
Prompt and image & at most 5,632 tokens \RCnext
Sampled transcription & at most 16,384 tokens \RCnext
Hardware & eight H200 GPUs per node \RClast
\bottomrule
\end{tabularx}
\caption{RLVR settings.}
\label{tab:rl-setup}
\end{table}

\Cref{tab:grpo-data} gives the composition of the RLVR pages and \Cref{tab:train-image-dpi} the resolution of their images.

\begin{table}[t]
\RCsetup
\begin{tabularx}{\textwidth}{>{\raggedright\arraybackslash}Xrr}
\toprule
\RChead{} & \RChead{Pages} & \RChead{Share} \\
\midrule
\RCgroup{Document type}{}
Synthetic pages from our generators & 3,788 & 33.7\% \RCnext
Curated table, chart and figure collections & 2,609 & 23.2\% \RCnext
Government and regulatory filings (FEC, IRS) & 1,799 & 16.0\% \RCnext
Public document-layout datasets (DocLayNet, RVL-CDIP) & 952 & 8.5\% \RCnext
Forms & 803 & 7.1\% \RCnext
Economic and central-bank publications (NBER, BIS, ECB, OECD) & 451 & 4.0\% \RCnext
Historical manuscripts and magazines & 258 & 2.3\% \RCnext
Financial filings & 123 & 1.1\% \RCnext
Contracts and redlines (CUAD) & 121 & 1.1\% \RCnext
Miscellaneous & 330 & 2.9\% \RClast
\RCgroup{How the reference transcription was made}{}
Teacher model, from the page image only & 5,613 & 50.0\% \RCnext
Generator (synthetic page) & 3,669 & 32.7\% \RCnext
Teacher model, anchored to the PDF text layer & 1,229 & 10.9\% \RCnext
PDF text layer only, no model & 723 & 6.4\% \RClast
\bottomrule
\end{tabularx}
\caption{The 11,234 pages of the training split, by what kind of document each page comes from and by how its reference transcription was made. Document types are inferred from the page identifier, so the groups are reliable and the individual counts approximate. Of the references, 6.4\% come from a page's own text layer and 32.7\% from the generator that drew the page; the rest were written by a model.}
\label{tab:grpo-data}
\end{table}

\begin{table}[t]
\RCsetup
\begin{minipage}{0.5\textwidth}
\begin{tabularx}{\linewidth}{>{\raggedright\arraybackslash}Xr}
\toprule
\RChead{Implied resolution} & \RChead{Share} \\
\midrule\RCstart
$\sim$200\,dpi & 56.2\% \RCnext
$\sim$150\,dpi & 28.7\% \RCnext
$\sim$100\,dpi & 7.4\% \RCnext
Square (non-page aspect) & 6.8\% \RCnext
Other ($\leq$75\,dpi) & 0.8\% \RCnext
Other ($\geq$225\,dpi) & 0.3\% \RClast
\bottomrule
\end{tabularx}
\end{minipage}
\caption{Page-image resolution in RLVR. The implied resolution is the long edge of the image divided by 11 inches.}
\label{tab:train-image-dpi}
\end{table}

\subsection{Training resolution}
\label{app:train-resolution}

Every training page is rendered at 1,024\,px on its long side, about 93\,dpi and 800 image tokens for a portrait letter page, while the benchmarks serve pages at a higher resolution (\Cref{evaluations}).
We also trained at a resolution closer to the serving one, and no transcription score changed by more than about a point.

We trained Qwen3.5-4B twice on a 30k-page subset of the SFT data, with the same recipe as the main runs, once with every page rendered at 1,024\,px and once at 1,536\,px, which gives 2.25 times the image tokens.
The 1,536\,px run took 1.2 times as long.
\Cref{tab:train-resolution} gives the scores, with every benchmark served at its usual rendering.
On ParseBench and PulseBench-Tab no score changed by more than 0.01, and on olmOCR-bench the table tests rose by 0.8 points and the math tests fell by 1.5.

The resolution at serving time matters far more than the resolution at training time.
We served the same two models on ParseBench in two ways: at the benchmark's standard rendering (150\,dpi, about 1,650\,px on the long side of a letter page) and with every page downscaled to at most 1,024\,px.
Downscaling cost both models about 0.08: the 1,536\,px model fell from 0.83 to 0.75 and the 1,024\,px model from 0.84 to 0.76.
A model trained only at 1,024\,px therefore still benefits from the extra image tokens of a larger page at serving time (\Cref{app:encoder}), and downscaling pages to save tokens is expensive.

\begin{table}[!htpb]
\RCsetup
\begin{minipage}{0.55\textwidth}
\begin{tabularx}{\linewidth}{>{\raggedright\arraybackslash}Xrr}
\toprule
\RChead{Score} & \RChead{1,024\,px} & \RChead{1,536\,px} \\
\midrule
\RCgroup{ParseBench}{}
Mean & 0.83 & 0.83 \RCnext
Table & 0.87 & 0.88 \RCnext
Chart & 0.79 & 0.79 \RCnext
Text content & 0.85 & 0.85 \RCnext
Text formatting & 0.80 & 0.80 \RClast
\RCgroup{olmOCR-bench}{}
Overall & 80.5 & 80.4 \RCnext
Table tests & 88.0 & 88.8 \RCnext
Math tests & 80.8 & 79.3 \RCnext
Reading order & 71.8 & 71.6 \RClast
\RCgroup{PulseBench-Tab}{}
Mean & 0.82 & 0.83 \RCnext
Perfect tables & 400 & 390 \RClast
\bottomrule
\end{tabularx}
\end{minipage}
\caption{Training at 1,024 against 1,536\,px on the same 30k pages. Qwen3.5-4B, plain transcription, two epochs at batch 64 and learning rate $1\times10^{-5}$; every benchmark served at its usual rendering.}
\label{tab:train-resolution}
\end{table}

\subsection{Reward components}
\label{app:reward-components}

\Cref{tab:reward-content,tab:reward-layout} list every component of the reward, what it scores, and the benchmark rule it mirrors where one exists.
The components are grouped by what they act on, and each is labelled with its panel, the group of components that shares one weight (\Cref{app:reward-formulas}).
The weights are a setting of each run, so the tables give the composition of the reward and not its weights.

\begin{table}[!htpb]
\RCsetup
\begin{tabularx}{\textwidth}{>{\raggedright\arraybackslash}p{3.3cm}>{\raggedright\arraybackslash}X>{\raggedright\arraybackslash}p{2.7cm}}
\toprule
\RChead{Component} & \RChead{What it scores} & \RChead{Mirrors} \\
\midrule
\RCgroup{Whole-page transcription}{panels \LIcode{bench}, \LIcode{text}}
\LIcode{bench} & benchmark pass rate on the completion, weighted by content type; presence-based, so it rewards recall at no precision cost & olmOCR-bench \RCnext
\LIcode{cct} & transcription precision, $1-\mathrm{lev}(p,g)/\max(|p|,|g|)$ on de-tabbed plaintext; penalises extra, reordered and marked-up text & edit distance, SCORE \citep{li2025score} \RCnext
\LIcode{eos} & clean termination & \RClast
\RCgroup{Tables}{panels \LIcode{table}, \LIcode{tbl_structure}}
\LIcode{table_align} & lead table term: mean of detection, cell-index accuracy and cell-content accuracy & SCORE \LIcode{score_tables}, by direct call \RCnext
\LIcode{table_teds} & tree-edit-distance similarity, from the same call, kept because both benchmarks report it & TEDS \citep{zhong2020pubtabnet} \RCnext
\LIcode{table_grits} & grid table similarity & GriTS \citep{smock2022grits} \RCnext
\LIcode{table_grits_trm} & ParseBench's own table-dimension metric & the table dimension \citep{zhang2026parsebenchdocumentparsingbenchmark} \RCnext
\LIcode{table_struct} & structure-only tree-edit distance, text-blind and span-aware, which catches the \LIcode{colspan} and \LIcode{rowspan} cases the content metrics read as duplicates; implemented behind a switch and not enabled in the runs reported here & TEDS structure variant \RCnext
\LIcode{tbl_structure} & structure agreement scored in a panel of its own, the only structural table signal carrying weight in the runs reported here & \RCnext
\LIcode{table_seg_penalty} & $1-\min(n_p,n_g)/\max(n_p,n_g)$ over table counts, so a merge and a split are penalised alike; a regex tag count, never on the per-cell path & \RClast
\RCgroup{Charts}{panel \LIcode{chart}}
\LIcode{chart_keying} & fraction of reference data cells placed under the same row and column headers & \LIcode{chart_data_point} \RCnext
\LIcode{chart_key_penalty} & penalty on a chart page that prints the reference numbers but leaves them unaddressable, counting only points whose failure reason is unassociated labels & \LIcode{chart_data_point}, by direct call \RCnext
\LIcode{chart_caption} & caption bound to its chart & \RClast
\RCgroup{Inline formatting and figure content}{panels \LIcode{formatting}, \LIcode{imgalt}, \LIcode{furniture}, \LIcode{diagram}}
\LIcode{formatting} & inline formatting marks & the formatting tests \RCnext
\LIcode{latex} & fraction of the reference's formulas reproduced, scored with the benchmark's own formula extractor & \LIcode{is_latex} \RCnext
\LIcode{imgalt} & image description content, graded & feature-bench alt axis \RCnext
\LIcode{furniture} & running heads, footers and page numbers lifted into their bands & \RCnext
\LIcode{diagram} & flowchart rendered as a fenced diagram block & \RClast
\bottomrule
\end{tabularx}
\caption{Reward components that score what the page says: the transcription itself, its tables and charts, and its inline formatting. Panel names are given with each group.}
\label{tab:reward-content}
\end{table}

\clearpage
\begin{table}[!htpb]
\RCsetup
\begin{tabularx}{\textwidth}{>{\raggedright\arraybackslash}p{3.3cm}>{\raggedright\arraybackslash}X>{\raggedright\arraybackslash}p{2.7cm}}
\toprule
\RChead{Component} & \RChead{What it scores} & \RChead{Mirrors} \\
\midrule
\RCgroup{Regions and boxes}{panels \LIcode{layout}, \LIcode{coverage}, \LIcode{binding}, \LIcode{emission}, \LIcode{bbox}}
\LIcode{layout} & F1 of expected against emitted region identifiers & layout element rule \RCnext
\LIcode{coverage} & recall of reference regions the output attributed, counting a region as attributed when its marker carries non-empty text; separates attributing a region from transcribing it well & layout element rule \RCnext
\LIcode{binding} & of the regions attributed, the fraction whose text is routed to the right identifier, by requiring that its bound text best-match that region's reference over all regions; targets right content bound to the wrong region & layout attribution \RCnext
\LIcode{grounding_bind} & mean soft-distance credit over emitted point markers & \RCnext
\LIcode{emission_penalty} & fraction of emitted region markers that are waste: an empty body on a text-bearing region, a repeated identifier, or one absent from the region table & \RCnext
\LIcode{bbox} & graded IoU F1 of emitted against reference boxes, gated on the label & box F1 at IoU 0.5 \RCnext
\LIcode{bbox_leak_penalty} & layout boxes emitted on a page whose prompt did not ask for them, the failure mode the two-prompt scheme of the SFT stage has to avoid & \RClast
\RCgroup{Output guards}{panels \LIcode{penalties}, \LIcode{structure}}
\LIcode{order} & reading-order fidelity, longest increasing subsequence over reference-ordered anchors & reading-order rule \RCnext
\LIcode{format_penalty} & on a table or chart page, requires an HTML table that parses to at least one real cell; a missing table, prose, a markdown table and an empty envelope fail equally & the detection gate \RCnext
\LIcode{length_penalty} & output running long against the reference past a 1.3$\times$ dead band & \RCnext
\LIcode{rep_penalty} & repeated n-gram fraction, against loops and runaway generation & \RCnext
\LIcode{marker_penalty} & marker hygiene & \RCnext
\LIcode{style_penalty} & one-sided penalty when styling exceeds the page's verified mark budget; the formatting tests are recall-only, so indiscriminate styling is otherwise profitable & \RCnext
\LIcode{double_emit_penalty} & the same content emitted twice & \RCnext
\LIcode{table_finish_penalty} & a page whose reference has a real table and whose prediction leaves a table unbalanced or far short of its rows; both legs are required, since the unbalanced leg alone makes omitting a table cheaper than truncating one & \RCnext
\LIcode{struct_penalty} & a degenerate-output guard, a flat penalty when the output runs past 20,000 characters at over 0.8 repetition or leaves runaway unclosed table tags; despite the name it is unrelated to table structure & \RClast
\bottomrule
\end{tabularx}
\caption{Reward components that score where content is placed and that guard against degenerate output. Panel names are given with each group.}
\label{tab:reward-layout}
\end{table}

\subsubsection{How the components are combined}
\label{app:reward-formulas}

The thirty-four components are grouped into sixteen panels, and the panels into nine targets.
The nine targets are a benchmark pass rate, content fidelity, table content, table structure, chart value-label keying, layout and attribution, layout boxes, formatting marks, and output hygiene.
Each component $k$ is either a score $s_k$ or a penalty $p_k$, both in $[0,1]$, and it contributes to the reward with its weight $w_k$:
\begin{equation}
c_k(y) =
\begin{cases}
\phantom{-}w_k\,s_k(y) & \text{if $k$ is a score,}\\
-w_k\,p_k(y) & \text{if $k$ is a penalty,}
\end{cases}
\label{eq:component}
\end{equation}
where a penalty is zero when the output does not have the error that it checks for.
The components of a panel $P_j$ are added, and the panels are combined with weights $\omega_j$ normalised to sum to one:
\begin{equation}
R(y) = \sum_{j=1}^{16}\frac{\omega_j}{\sum_{l}\omega_l}\sum_{k\in P_j} g_k(y)\,c_k(y).
\label{eq:reward}
\end{equation}
A length gate multiplies the seven content scores, such as \LIcode{cct} and the table scores, and removes or reduces their credit when the output is much longer or much shorter than the reference or repeats itself.
The \LIcode{length_penalty} also lowers the reward directly when the output is too long.
For short outputs we use only the gate, because a penalty for short outputs can give an empty output a higher reward than an output with half of the page.

In \Cref{eq:reward} the gate is the factor $g_k$.
Let $n$ and $n^{*}$ be the lengths in characters of the output and of the reference.
For \LIcode{bench}, \LIcode{cct}, \LIcode{table_align}, \LIcode{table_teds}, \LIcode{table_grits}, \LIcode{table_grits_trm} and \LIcode{chart_keying}, $g_k$ is the product $g_{\mathrm{over}}\,g_{\mathrm{under}}\,g_{\mathrm{rep}}$, and for every other component it is one.
The first factor removes the credit of an output that is much longer than the reference, and the second reduces the credit of an output that is much shorter:
\begin{align}
g_{\mathrm{over}} &=
\begin{cases}
1 & \rho \le 1.5,\\
(4-\rho)/2.5 & 1.5 < \rho < 4,\\
0 & \rho \ge 4,
\end{cases}
& \rho &= \frac{n}{n^{*}+250/1.5},
\label{eq:gate-over}\\[4pt]
g_{\mathrm{under}} &=
\begin{cases}
1 & \rho \ge 0.6,\\
0.3+0.7\,(\rho-0.2)/0.4 & 0.2 < \rho < 0.6,\\
0.3 & \rho \le 0.2,
\end{cases}
& \rho &= \frac{n+250}{n^{*}}.
\label{eq:gate-under}
\end{align}
The third factor is $g_{\mathrm{rep}} = 1-f_{\mathrm{rep}}$, where $f_{\mathrm{rep}}$ is the fraction of the raw output that is a runaway repetition.
The 250 characters added in both ratios prevent a short page from being gated for a small absolute difference in length.
The under-generation gate has a floor of 0.3.
The gates, like the weights, are set per run.

The \LIcode{length_penalty} component is a penalty for the same error as $g_{\mathrm{over}}$.
It is zero while the output is within its allowed length, and then it increases linearly to one:
\begin{equation}
p_{\mathrm{len}} = \min\!\left(1,\ \max\!\left(0,\ \frac{n-\max(1.3\,n^{*},\ n^{*}+250)}{\max(n^{*},\ 250)}\right)\right).
\label{eq:length-penalty}
\end{equation}

\subsubsection{Tests for reward terms}
\label{app:reward-tests}

Every reward term is tested on constructed answers before training.

The tests led to three rules for writing a reward term.
A component with nothing to compare must return zero, because returning a fixed middle score instead gave an empty answer a quarter of the summed score.
A check on order must require content, because an empty sequence is trivially in order and would otherwise score well on reading order.
A check on counts or labels must require correct placement, because otherwise an answer with every element in the wrong place still scores well.

\subsubsection{How the reward compares}
\label{app:reward-comparison}

The document-parsing VLMs that add a reinforcement-learning stage use similar algorithms and differ mainly in what their rewards score.
\Cref{tab:reward-comparison} places ours beside them, using each system's own account of its reward.
Beyond covering more of the page, our reward differs in how finely it scores and in its penalties.

An edit-distance reward is smooth, but it is computed over the whole page and cannot say which cell of a table is wrong.
A binary unit test says exactly what it checks and gives nothing in between, so a nearly correct table and an absent one are worth the same.
Infinity-Parser2 grades cropped tables with tree edit distance, but it scores a full page with an edit distance.
Our terms are graded where a partial parse deserves partial credit, and binary where a page either carries a parseable table or does not.

Twelve of our thirty-four components are penalties.
A reward built only from scores costs an output nothing for repeating itself, running past the end of the page or carrying formatting the page does not have.
Two of the published rewards guard against repetition.
olmOCR~2 rewards an output that ends with the end-of-sequence token, and LightOnOCR penalises repetition loops.
None of them describes penalties for split or merged tables, unfinished tables or formatting the page does not have.
Our reward needs the formatting penalty because the formatting tests measure recall alone, so without it, styling every line raises the score.

\begin{table}[!htpb]
\RCsetup
\begin{tabularx}{\textwidth}{>{\raggedright\arraybackslash}p{3.1cm}>{\raggedright\arraybackslash}X>{\raggedright\arraybackslash}p{2.3cm}>{\raggedright\arraybackslash}p{2.1cm}}
\toprule
\RChead{System} & \RChead{What the reward scores} & \RChead{Granularity} & \RChead{Penalties} \\
\midrule
\RCgroup{Published models}{as their authors describe them}
Infinity-Parser \citep{wang2025infinityparserlayoutaware} & edit distance, paragraph count, reading order & whole page & none described \RCnext
olmOCR~2 \citep{poznanski2025olmocr2unittest} & binary unit tests, on synthetic pages whose source markup gives the answer; an output that ends with the end-of-sequence token; document metadata & per test, binary & none described; the end-of-sequence reward guards against outputs that never stop \RCnext
LightOnOCR \citep{taghadouini2026lightonocr1bendtoendmultilingual} & the olmOCR~2 unit tests and formulas that render; in a separate variant, the overlap of the emitted image boxes with the true boxes & per test; per box & repetition loops, malformed formulas, missing and invented boxes \RCnext
Infinity-Parser2 \citep{huang2026infinityparser2technicalreport} & eight tasks jointly, each with its own metric: tree edit distance for cropped tables, value-and-label matching for charts converted to tables, edit distance for full pages & per task & none described \RClast
\RCgroup{This work}{}
Ours & nine targets, from the transcription itself to layout attribution and output hygiene & per component, graded and binary & twelve of thirty-four components \RClast
\bottomrule
\end{tabularx}
\caption{The reward of each reinforcement-trained document-parsing VLM, as its authors describe it. ``None described'' records that the published account does not mention penalties against degenerate output, not that the system has none.}
\label{tab:reward-comparison}
\end{table}

\subsubsection{Combining the terms}
\label{app:reward-combining}

\paragraph{Variance reweighting.}
\ocria{} is trained with hand-tuned panel weights, with variance reweighting as a safeguard.
Let $\sigma_j$ be the standard deviation of panel $j$ within a group, averaged over the groups of a step.
We keep a running estimate $S_j$ and, after 15 warm-up steps, scale the panel weight by it:
\begin{equation}
S_j \leftarrow 0.9\,S_j + 0.1\,\sigma_j,
\qquad
\omega_j \leftarrow \omega_j\cdot\min\!\bigl(1,\ \max(0.3,\ S_j/0.05)\bigr).
\label{eq:variance-reweighting}
\end{equation}
A panel whose rollouts no longer differ loses up to 70\% of its weight.
A panel can be exempted when its low variance means that the model is stuck on it and has not mastered it.

\paragraph{Balancing the terms.}
In a group-relative update, the influence of a term depends on how much it varies within the group as well as on its weight.
A term with a small weight and a large variance can therefore dominate the gradient, while a term that barely varies, such as the content term, rarely decides the ranking.
In one training run, the formatting term became the largest single contribution to the ranking within ten steps, the contribution of the table term fell by half, and the table scores regressed.
Normalising the panel weights to sum to one also lowers the weight of every other panel when one target is split into two panels, and giving the formula term its own panel cut its effective weight by a factor of four.
The published remedy is to normalise each component by its own within-group statistics before summing \citep{mogrpo2025}, but it did not work for our reward.
Several of our components are one-sided guardrails, penalties that are zero when nothing is wrong, and normalising them to unit variance gives them as much influence as the terms they protect; on stored groups, it moved a share of the gradient onto them.
In a training run with this normalisation, the policy removed formatting markup to avoid the formatting penalty, the formatting score collapsed and the overall score fell with it.
A revised normalisation leaves the guardrails out of the normalised set, puts a floor on the within-group standard deviation, clips the normalised score and blends the result with the ordinary advantage of the weighted reward.
On the same stored groups, it removes the guardrails' share of the gradient and raises the share of the chart and table terms, and Constrained GRPO \citep{girgis2026cgrpo} independently argues for combining standardised advantages over standardising a scalarised reward.
We have tested it only offline, and \ocria{} is not trained with it.
Reweighting moves gradient between objectives but cannot add signal, so a chart page on which every rollout fails in the same way gives no signal with any weights.
Because the gradients of the objectives are correlated, projecting one gradient out of another \citep{yu2020pcgrad} also changes the others.

\subsection{Pages with no training signal}
\label{app:page-selection}

Published remedies for pages whose rollouts agree are to filter batches by disagreement \citep{yu2025dapo}, to select prompts by their effect on learning \citep{limr2025}, and to schedule difficulty as the model improves \citep{shi2025adarft,chen2025sec}.
We report results for two of them.

\paragraph{Filtering during training.}
We filtered batches by disagreement on one objective.
This increased that objective's share of the gradient but reduced the signal for the other objectives.
The filter reads the current policy's own disagreement, which changes as the policy changes.

\paragraph{Selecting the pages before training.}
This test removes, before training starts, the pages that the SFT model is unlikely to learn from.
We sampled each candidate page eight times with the SFT model and set aside the 6,096 pages whose samples agreed.
The SFT model had already transcribed 4,713 of them correctly and was uniformly poor on 1,215.
On the remaining 168, the agreement came from a suspect reference.
A page that the SFT model already transcribes is unlikely to divide the policy trained from it, so these pages are the least likely to carry signal later.
Training on the remaining pages stayed ahead of the full corpus at both steps we measured, by a margin that narrowed, and the two runs improved on different dimensions.
\ocria{} is trained on the full corpus, because removing these pages gave mixed results (\Cref{rl}), and the corpus still contains 3,473 of them.

\section{Encoder experiments}

This appendix studies where small text is lost between the page image and the transcription, and how the number of image tokens per page changes the scores.
\label{app:encoder}

The encoder of Qwen cuts a page image into $16{\times}16$\,px patches, and the projector joins each $2{\times}2$ group of patches into one image token, so one image token covers $32{\times}32$\,px of the page.
At 150\,dpi that is three to four characters of 10\,pt text and about twenty characters of 4\,pt text.
We measured each stage between the page and the transcription separately, all on the Qwen3.5-4B model.
A probe is a small classifier trained to name the character at a position from the features at that position.
Its accuracy says how readable the character is at that stage, not whether the information is gone.

\begin{table}[!htpb]
\RCsetup
\begin{tabularx}{\textwidth}{>{\raggedright\arraybackslash}p{2.0cm}>{\raggedright\arraybackslash}p{4.6cm}>{\raggedright\arraybackslash}X}
\toprule
\RChead{Stage} & \RChead{Measurement} & \RChead{Result} \\
\midrule\RCstart
Rendering & image tokens per table cell, and type height & 63 tokens per cell and 20\,px type on pages up to 100 cells; 1.1 tokens per cell and 8\,px above 1,000 cells \RCnext
Vision encoder & probe accuracy for character identity as the page gets fewer image tokens & 0.82 at 3,796 tokens, 0.40 at 1,024, 0.22 at 400, 0.11 at 144 (chance floor about 0.09) \RCnext
Projector & probe accuracy before and after the projector & digits 0.84 $\rightarrow$ 0.76, letters 0.56 $\rightarrow$ 0.41; gutters, column boundaries and merged cells drop by at most 0.02 \RCnext
Decoder & activation patching on degraded pages & clean image tokens at the decoder input restore all of the lost quality; at layer 16 they restore 50\% (95\% CI 39--63\%) \RClast
\bottomrule
\end{tabularx}
\caption{Where small text is lost, stage by stage, on the Qwen3.5-4B model.}
\label{tab:encoder-stages}
\end{table}

Small text is weakened before the decoder sees it, mostly by the resolution the encoder receives and further by the projector.
Table structure crosses the projector almost unchanged, and the decoder reads what it is given.
Fine-tuning changed the vision side little: character-identity probes before and after the projector are within 0.03 of the base model's.
Inside the decoder it changed more, with probes at layer 16 up to 0.06 above the base model's.

\begin{table}[!htpb]
\RCsetup
\begin{tabularx}{\textwidth}{>{\raggedright\arraybackslash}Xrrrrrr}
\toprule
\RChead{Cells} & \RChead{Pages} & \RChead{\shortstack[r]{Table\\score}} & \RChead{\shortstack[r]{Image tokens\\per cell}} & \RChead{\shortstack[r]{Type\\height (px)}} & \RChead{\shortstack[r]{Reference\\length (tokens)}} & \RChead{\shortstack[r]{Abnormal\\end}} \\
\midrule\RCstart
1--100 & 259 & 0.96 & 63.0 & 19.6 & 596 & 0 \RCnext
101--300 & 142 & 0.92 & 12.6 & 18.6 & 2,368 & 0 \RCnext
301--600 & 61 & 0.82 & 5.0 & 14.6 & 5,111 & 1 \RCnext
601--1,000 & 28 & 0.88 & 3.2 & 13.2 & 9,196 & 0 \RCnext
Over 1,000 & 13 & 0.59 & 1.1 & 8.0 & 24,139 & 6 \RClast
\bottomrule
\end{tabularx}
\caption{ParseBench table pages by size (4B model, 150\,dpi, output limit 24,576 tokens). The table score is the mean over pages; the other columns are medians.}
\label{tab:dense-tables}
\end{table}

\paragraph{More image tokens help and help fine text most.}
\Cref{tab:token-budget} serves olmOCR-bench to the same weights with different numbers of image tokens.
Going from 1,240 to 2,268 tokens per page by enlarging the same 1,288\,px render takes long tiny text from 92.3\% to 93.9\% and the overall score from 83.5\% to 84.4\%.
Those extra tokens carry no new detail, and rendering the page at 1,822\,px instead, for 2,451 tokens, adds only 0.3 points more on long tiny text, though 1.5 points more overall.
For fine text, what limits the model is how many tokens each glyph gets, not the detail in the pixels.
The model was trained on pages rendered at 1,024\,px, about 800 image tokens, so every standard evaluation rendering here gives it more tokens per glyph than it saw in training.

\begin{table}[!htpb]
\RCsetup
\begin{minipage}{0.75\textwidth}
\begin{tabularx}{\linewidth}{r>{\raggedright\arraybackslash}Xrr}
\toprule
\RChead{Image tokens} & \RChead{Page image} & \RChead{olmOCR-bench} & \RChead{Long tiny text} \\
\midrule\RCstart
266 & 1,288\,px render (shrunk) & 45.7 & 7.9 \RCnext
532 & 1,288\,px render (shrunk) & 67.2 & 36.9 \RCnext
1,240 & 1,288\,px render & 83.5 & 92.3 \RCnext
2,268 & 1,288\,px render (enlarged) & 84.4 & 93.9 \RCnext
2,451 & 1,822\,px render & 85.9 & 94.2 \RClast
\bottomrule
\end{tabularx}
\end{minipage}
\caption{olmOCR-bench for the 4B model as the image-token budget changes. Image tokens are the median per page.}
\label{tab:token-budget}
\end{table}

\paragraph{\texorpdfstring{Retraining the projector alone did not help.}{Retraining the projector alone did not help.}}
{On the base model, fine-tuning only the projector raised probe accuracy by 0.08 and dropped olmOCR-bench from 71.0 to 48.0, so we did not adopt it.}

\paragraph{\texorpdfstring{Scope.}{Scope.}}
{Every benchmark number is a single run with greedy decoding.
Every measurement is on the 4B model except the projector retraining, which was run on the base model.}
\section{Evaluation details}
\label{app:eval-protocol}

This appendix gives the protocol of the evaluation in \Cref{evaluations}, the results for each benchmark (\Cref{app:eval-breakdown}) and the cases where the scores are lowest (\Cref{eval:failures}).

\paragraph{Transcription benchmarks.}
ParseBench~\citep{zhang2026parsebenchdocumentparsingbenchmark} grades a page on text content, text formatting, tables and charts.
Text content, text formatting and charts are scored as the pass rate of rules written against the page, and tables with the combination of GriTS and TableRecordMatch (GTRM); we report the mean of the four.
Pages are rendered at 300\,dpi, and the model's image processor scales a letter page down to about 4.1 megapixels, about 210\,dpi.
olmOCR-bench~\citep{poznanski2025olmocrunlockingtrillionstokens} holds 1,403 pages and 7,010 unit tests (a string present or absent, two strings in order, a cell under its header, an equation rendering to the same symbols), rendered at 1,288\,px on the long side and scored as the mean over its seven categories and the baseline test.
PulseBench-Tab~\citep{pulsebenchtab2026} holds 1,820 cropped tables in nine languages scored by the T-LAG grid metric. We send the crop capped at 2,560\,px with a 16k-token budget, score the largest HTML table in the answer, and report the share of tables scored, since the protocol leaves tables without a parseable prediction out of the mean.
fr-bench-pdf2md~\citep{frbenchpdf2md2026} holds 721 difficult French pages with 2,318 olmOCR-style tests, scored as the pass rate over all of them. We render its pages at 2,048\,px on the long side.

\paragraph{Post-processing.}
The model is trained on our output format (\Cref{appendix:annotations}), which includes constructs that these benchmarks neither expect nor score, such as page headers and footers inside their own tags, a watermark tag, an image marker with a label, a Mermaid block for a diagram, and a horizontal rule for a printed one.
Before scoring we remove these constructs from every candidate and normalize the markup inside table cells (emphasis, checkboxes, and superscripts and subscripts written back as the glyphs the page prints).

\paragraph{Rendering resolution.}
We serve each benchmark at its own standard rendering: olmOCR-bench pages at 1,288\,px on the long side (about 1,240 image tokens for a letter page), and ParseBench pages at 300\,dpi, which the image processor scales down to about 3,900 image tokens for a letter page.
Both are above the 1,024\,px, about 800 image tokens, that the model is trained on.
Scores depend on this choice.
With no change to the weights, giving the \ocria{} model 2,451 image tokens per page instead of 1,240 raises its olmOCR-bench score by 2.4 points (\Cref{app:encoder}).
Other models respond the same way. Chandra-OCR-2 and LightOnOCR-3 improve by 2 points on olmOCR-bench when their pages are rendered at 212\,dpi.
\begin{LINote}
A difference of 2--3 points between two systems can come from rendering alone, so we use each benchmark's standard rendering for the systems we evaluate ourselves.
On olmOCR-bench the other systems scored much lower at our rendering, so we report their published numbers and run \ocria{} at the standard 1,288\,px.
\Cref{tab:olmocr-same-render} combines our runs at 1,288\,px with published scores for Mistral OCR 4.1 and olmOCR-2.
\end{LINote}

\paragraph{Other systems.}
On ParseBench, PulseBench-Tab and fr-bench-pdf2md every other system is run by us at the same renderings and scored by the same scorers.
All open models (Infinity-Parser2-Pro, LightOnOCR-3-4B, Chandra-OCR-2, dots.mocr, LightOnOCR-2-1B) are served with vLLM at each model's own image budget.
LightOnOCR-3-4B runs in its grounding mode, which scores higher than its plain transcription mode on tables and charts (0.75 against 0.61 on ParseBench).
olmOCR-2 runs through its own pipeline, which renders the pages itself at the size we pass.
Surya OCR 2 runs through its own package, on vLLM.
Where a system labels page headers, footers and pictures, those blocks are dropped the way our own tags are, and pipe tables are converted to HTML where a scorer wants HTML.

\subsection{Results by benchmark}
\label{app:eval-breakdown}

\Cref{tab:olmocr-categories,tab:pulse,tab:frbench} break down the results of \Cref{tab:main-results} for olmOCR-bench, PulseBench-Tab and fr-bench-pdf2md.

\Cref{tab:main-results} reports the mean of the Tables, arXiv math and Old scans math categories, the parts of olmOCR-bench that test tables and formulas.
We report this part because tables and formulas are the elements of the page that our training data targets (\Cref{data}), while olmOCR-bench has no test of the other two elements we focus on, charts and text formatting.
\Cref{tab:olmocr-categories} gives all eight categories and the official score, with the numbers that the other systems' authors publish.
\Cref{tab:olmocr-same-render} reports our runs alongside two published baselines.
For our runs, pages are rendered at 1,288\,px on the long side, the benchmark's standard rendering, and models are served with vLLM at their own image budgets.
We apply the same post-processing and the olmOCR-bench scorer to all outputs from these runs.
Mistral OCR 4.1's scores come from the LightOnOCR-3 post, and olmOCR-2's from its model card, which reports the same 1,288\,px rendering.
In this comparison \ocria{} has the highest official score, 83.5 against 83.2 for Infinity-Parser2-Pro, and the highest mean on the three table and formula categories, 88.6 against 87.0 for LightOnOCR-2-1B.
The six systems we served ourselves score 1.8 to 5.7 points lower here than in their published numbers, which is why the main table uses the published ones.

\begin{table}[t]
\RCsetup\scriptsize\setlength{\tabcolsep}{3pt}
\begin{tabularx}{\linewidth}{Xccccccccc}
\toprule
\RChead{Model} & \RChead{arXiv math} & \RChead{Old scans math} & \RChead{Tables} & \RChead{Old scans} & \RChead{Headers, footers} & \RChead{Multi-column} & \RChead{Long tiny text} & \RChead{Base} & \RChead{Overall} \\
\midrule\RCstart
Infinity-Parser2-Pro & 88.1 & \textbf{91.3} & 91.2 & \textbf{58.2} & \textbf{95.8} & 83.7 & 92.5 & \textbf{99.9} & \textbf{87.6} \RCnext
LightOnOCR-3-4B & \textbf{91.1} & 89.1 & 90.2 & 52.5 & 89.2 & \textbf{85.6} & 93.2 & 99.5 & 86.3 \RCnext
olmOCR-2 (7B) & 82.9 & 82.1 & 84.3 & 48.3 & 95.7 & 84.3 & 81.4 & 99.7 & 82.3 \RCnext
Chandra-OCR-2 (4B) & 86.9 & 89.1 & 92.1 & 51.1 & 91.4 & 82.1 & \textbf{93.7} & \textbf{99.9} & 85.8 \RCnext
dots.mocr (3B) & 85.9 & 85.5 & 90.7 & 48.2 & 94.0 & 85.3 & 81.6 & 99.7 & 83.9 \RCnext
Surya OCR 2 (0.7B) & 88.3 & 81.4 & 86.6 & 41.8 & 92.5 & 82.4 & \textbf{93.7} & 99.7 & 83.3 \RCnext
LightOnOCR-2-1B & 89.6 & 85.6 & 89.0 & 42.2 & 19.7 & 84.8 & 91.4 & 99.6 & 75.2 \RCnext
Mistral OCR 4.1 & 82.6 & 72.3 & 85.9 & 47.9 & 91.4 & 84.5 & 91.2 & 99.5 & 81.9 \RClast
\midrule\RCstart
\ocria{} & 89.6 & 89.1 & \textbf{92.9} & 48.9 & 81.4 & 81.6 & \textbf{93.7} & \textbf{99.9} & 84.6 \RClast
\bottomrule
\end{tabularx}
\caption{olmOCR-bench by category. Overall is the official average over the eight columns. The other systems' rows are the numbers they publish (the LightOnOCR-3 post; olmOCR-2's model card); the \ocria{} row is our run. LightOnOCR-2-1B's paper reports 83.2 with the headers-and-footers category left out; here its overall is recomputed over all eight categories, using the headers-and-footers score (19.7) from the same paper.}
\label{tab:olmocr-categories}
\end{table}

\begin{table}[t]
\RCsetup\scriptsize\setlength{\tabcolsep}{3pt}
\begin{tabularx}{\linewidth}{Xccccccccc}
\toprule
\RChead{Model} & \RChead{arXiv math} & \RChead{Old scans math} & \RChead{Tables} & \RChead{Old scans} & \RChead{Headers, footers} & \RChead{Multi-column} & \RChead{Long tiny text} & \RChead{Base} & \RChead{Overall} \\
\midrule\RCstart
Infinity-Parser2-Pro & 82.9 & 85.2 & 87.7 & \textbf{51.0} & 92.1 & 82.8 & 84.4 & \textbf{99.9} & 83.2 \RCnext
LightOnOCR-3-4B & 86.9 & 78.8 & 87.6 & 48.7 & 87.6 & 84.3 & 83.5 & 99.7 & 82.1 \RCnext
olmOCR-2 (7B) & 82.9 & 82.1 & 84.3 & 48.3 & \textbf{95.7} & 84.3 & 81.4 & 99.7 & 82.3 \RCnext
Chandra-OCR-2 (4B) & 80.3 & 84.3 & 86.7 & 45.8 & 90.8 & 80.7 & 81.2 & 99.9 & 81.2 \RCnext
dots.mocr (3B) & 77.9 & 78.4 & 86.3 & 40.9 & 90.0 & 82.2 & 85.3 & 99.6 & 80.1 \RCnext
Surya OCR 2 (0.7B) & 78.8 & 76.0 & 82.1 & 38.2 & 93.2 & 79.9 & 73.3 & 99.7 & 77.6 \RCnext
LightOnOCR-2-1B & \textbf{88.1} & 87.3 & 85.7 & 40.3 & 22.4 & 83.8 & 80.3 & 99.5 & 73.4 \RCnext
Mistral OCR 4.1 & 82.6 & 72.3 & 85.9 & 47.9 & 91.4 & \textbf{84.5} & 91.2 & 99.5 & 81.9 \RClast
\midrule\RCstart
\ocria{} & 85.0 & \textbf{87.8} & \textbf{92.9} & 46.6 & 81.8 & 81.4 & \textbf{92.3} & 99.8 & \textbf{83.5} \RClast
\bottomrule
\end{tabularx}
\caption{olmOCR-bench: our runs at 1,288\,px alongside published baselines. For our runs, pages are rendered at 1,288\,px on the long side, models are served with vLLM at their own image budgets, and outputs receive the same post-processing and are scored by the olmOCR-bench scorer. Mistral OCR 4.1's row is from the LightOnOCR-3 post and olmOCR-2's from its model card at the same 1,288\,px. Overall is the official average over the eight columns. Of the 440 arXiv-math tests that \ocria{} fails, 143 render to the same formula as the reference but write the \LaTeX{} in a form other than the KaTeX one the test expects; counting them as passed gives 89.9 on that category.}
\label{tab:olmocr-same-render}
\end{table}

\begin{table}[t]
\RCsetup
\begin{minipage}{0.72\linewidth}
\begin{tabularx}{\linewidth}{Xcccc}
\toprule
\RChead{Model} & \RChead{Mean} & \RChead{Coverage \%} & \RChead{Median} & \RChead{Perfect tables} \\
\midrule\RCstart
Infinity-Parser2-Pro & 0.82 & 96.8 & 0.93 & 419 (23.0\%) \RCnext
LightOnOCR-3-4B & 0.82 & 96.6 & 0.93 & 431 (23.7\%) \RCnext
olmOCR-2 (7B) & 0.80 & 92.4 & 0.90 & 354 (19.5\%) \RCnext
Chandra-OCR-2 (4B) & 0.80 & 93.0 & 0.92 & 340 (18.7\%) \RCnext
dots.mocr (3B) & 0.82 & 88.7 & 0.92 & 308 (16.9\%) \RCnext
Surya OCR 2 (0.7B) & 0.71 & 93.7 & 0.82 & 224 (12.3\%) \RCnext
LightOnOCR-2-1B & 0.81 & 89.6 & 0.90 & 354 (19.5\%) \RCnext
Mistral OCR 4.1 & 0.75 & 97.7 & 0.82 & 269 (14.8\%) \RClast
\midrule\RCstart
\ocria{} & \textbf{0.84} & \textbf{99.1} & \textbf{0.95} & \textbf{485 (26.6\%)} \RClast
\bottomrule
\end{tabularx}
\end{minipage}
\caption{Results on PulseBench-Tab, 1,820 tables. Mean and median T-LAG score over the tables scored; coverage is the share of tables with a parseable prediction.}
\label{tab:pulse}
\end{table}

\begin{table}[t]
\RCsetup\scriptsize\setlength{\tabcolsep}{3pt}
\begin{tabularx}{\linewidth}{Xcccccccc}
\toprule
& \multicolumn{8}{c}{\RChead{fr-bench-pdf2md, pass rate by category}} \\
\cmidrule(lr){2-9}
\RChead{Model} & \RChead{Forms} & \RChead{Graphics} & \RChead{Handwritten} & \RChead{Long table} & \RChead{Long table 2} & \RChead{Multi-column} & \RChead{Tiny text} & \RChead{Overall} \\
\midrule\RCstart
Infinity-Parser2-Pro & 33.9 & 26.9 & 33.3 & 51.5 & 50.0 & 76.4 & 78.7 & 54.9 \RCnext
LightOnOCR-3-4B & \textbf{49.6} & 68.3 & \textbf{41.2} & 80.6 & 70.7 & 83.6 & \textbf{83.2} & 71.4 \RCnext
olmOCR-2 (7B) & 43.8 & 37.7 & 20.6 & 67.9 & 63.9 & \textbf{84.8} & 66.2 & 56.9 \RCnext
Chandra-OCR-2 (4B) & 38.3 & 72.8 & 22.4 & 80.6 & 77.8 & 70.9 & 70.0 & 65.5 \RCnext
dots.mocr (3B) & 40.1 & 26.1 & 13.9 & 56.7 & 59.8 & 80.6 & 79.3 & 56.8 \RCnext
Surya OCR 2 (0.7B) & 22.6 & 26.6 & 1.8 & 74.6 & 66.1 & 73.3 & 64.5 & 50.9 \RCnext
LightOnOCR-2-1B & 35.4 & 32.9 & 12.1 & 73.9 & 61.3 & 78.8 & 64.7 & 53.5 \RCnext
Mistral OCR 4.1 & 39.4 & 55.8 & 11.5 & 69.4 & 59.8 & 73.3 & 57.4 & 54.1 \RClast
\midrule\RCstart
\ocria{} & 46.0 & \textbf{78.5} & 21.2 & \textbf{83.2} & \textbf{81.9} & 77.0 & 78.9 & \textbf{72.1} \RClast
\bottomrule
\end{tabularx}
\caption{Results on fr-bench-pdf2md, pass rate by category and overall, over 2,318 tests.}
\label{tab:frbench}
\end{table}

\subsection{Areas for improvement}
\label{eval:failures}

The scores are lowest on the hardest cases within the areas we worked on.
Old scans are the weakest olmOCR-bench category, 48.9\% for \ocria{} against 81.4 to 99.9\% for the other seven categories.
The other systems in \Cref{tab:olmocr-categories} score 41.8 to 58.2\% on old scans, so these pages remain hard for current document-parsing VLMs.
Tables in Arabic, Greek and Korean score 0.70 to 0.77 on PulseBench-Tab, against 0.86 to 0.90 for French, Chinese, Spanish, Russian and English.
Text formatting is the lowest of the four ParseBench dimensions for \ocria{}, although its 0.83 is the highest score on that dimension in \Cref{tab:parsebench-dims}.
Within it, bold passes 77\% of its rules and italics 63\%.
Highlighting passes 35\% and strikethrough 45\%, and both are rare on real pages.

\begin{figure}[!htbp]
\centering\setlength{\fboxsep}{0pt}%
\begin{minipage}[t]{0.49\textwidth}\vspace{0pt}\centering
\fcolorbox{LIPanelLine}{white}{\includegraphics[width=\dimexpr\linewidth-0.8pt\relax]{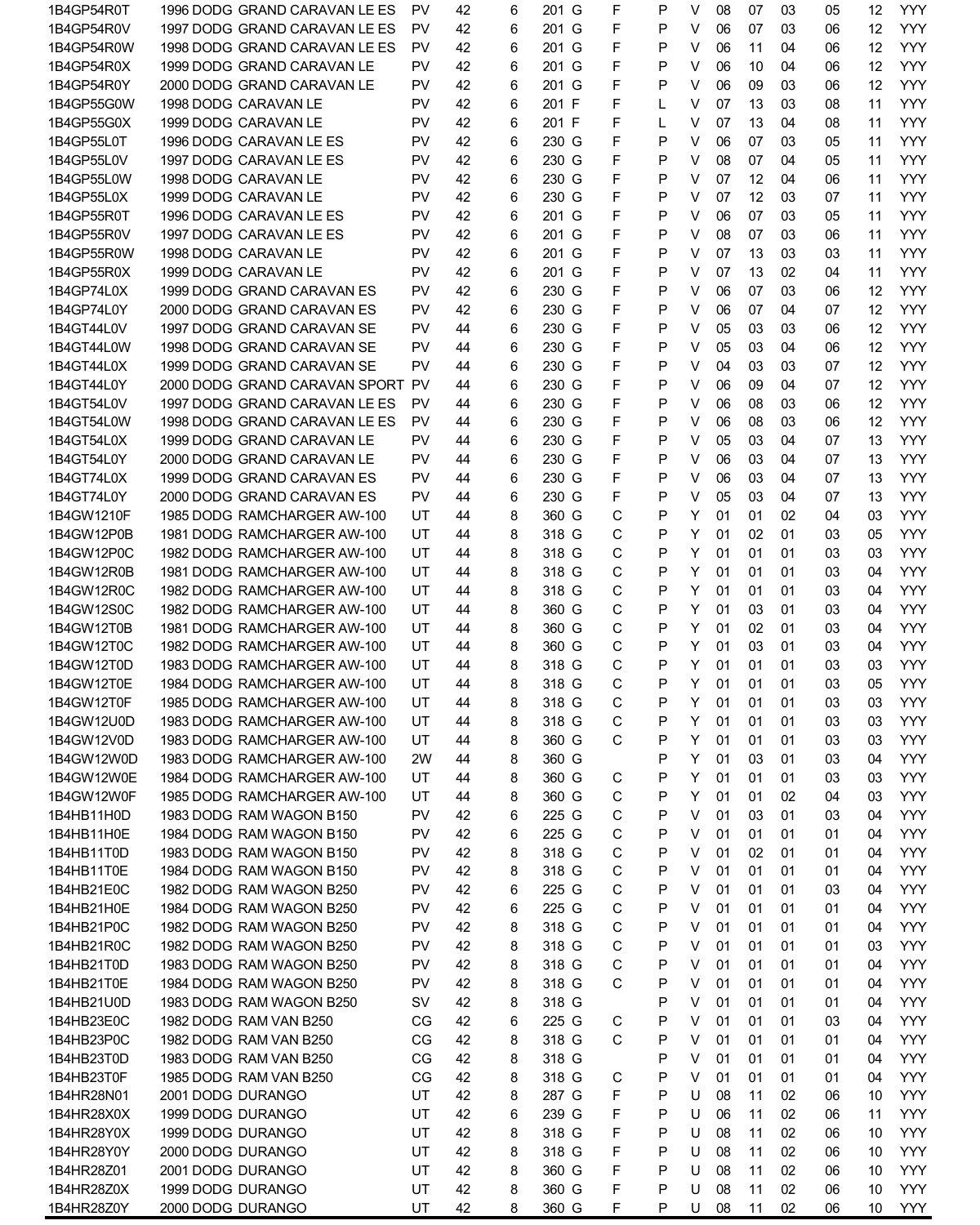}}\\[3pt]
{\scriptsize\sffamily\color{LIMuted}Page image (bottom)}
\end{minipage}\hfill
\begin{minipage}[t]{0.49\textwidth}\vspace{0pt}\centering
\fcolorbox{LIPanelLine}{white}{\includegraphics[width=\dimexpr\linewidth-0.8pt\relax]{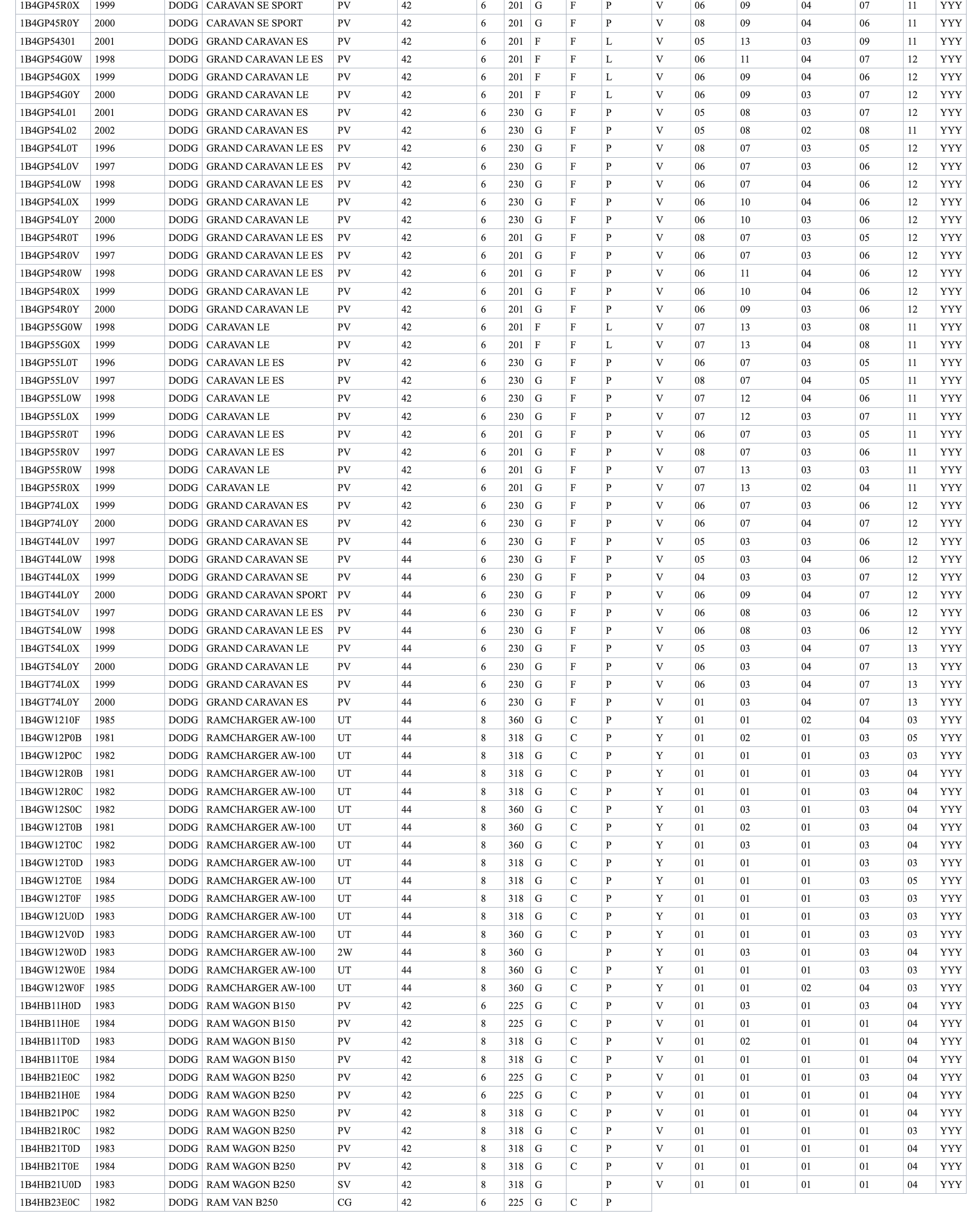}}\\[3pt]
{\scriptsize\sffamily\color{LIMuted}\ocria{} output, rendered (end)}
\end{minipage}
\caption{\textbf{A table longer than the output budget.} A page that holds one table of vehicle codes, 195 rows and 18 columns, about 3,500 cells. \ocria{} keeps the 18 columns and writes the rows in order, without repeating any, until it reaches the 24,576-token output limit after 183 rows. The last 12 rows are missing and the table has no closing tags. The left panel shows the bottom of the page and the right panel the end of the output.}
\label{fig:limit-long-table}
\end{figure}

\begin{figure}[!htbp]
\centering\setlength{\fboxsep}{0pt}%
\begin{minipage}[t]{0.49\textwidth}\vspace{0pt}\centering
\fcolorbox{LIPanelLine}{white}{\includegraphics[width=\dimexpr\linewidth-0.8pt\relax]{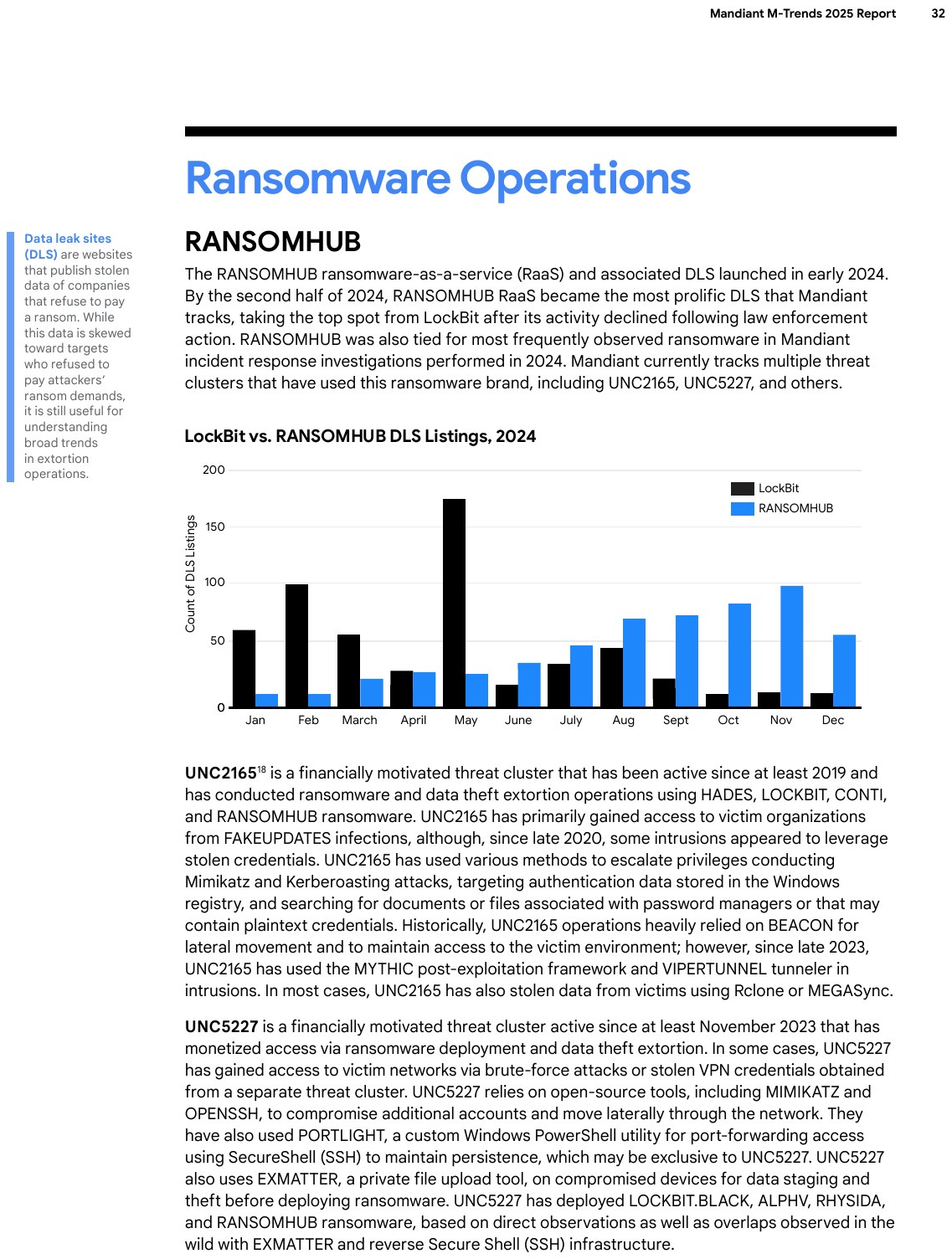}}\\[3pt]
{\scriptsize\sffamily\color{LIMuted}Page image}
\end{minipage}\hfill
\begin{minipage}[t]{0.49\textwidth}\vspace{0pt}\centering
\fcolorbox{LIPanelLine}{white}{\includegraphics[width=\dimexpr\linewidth-0.8pt\relax]{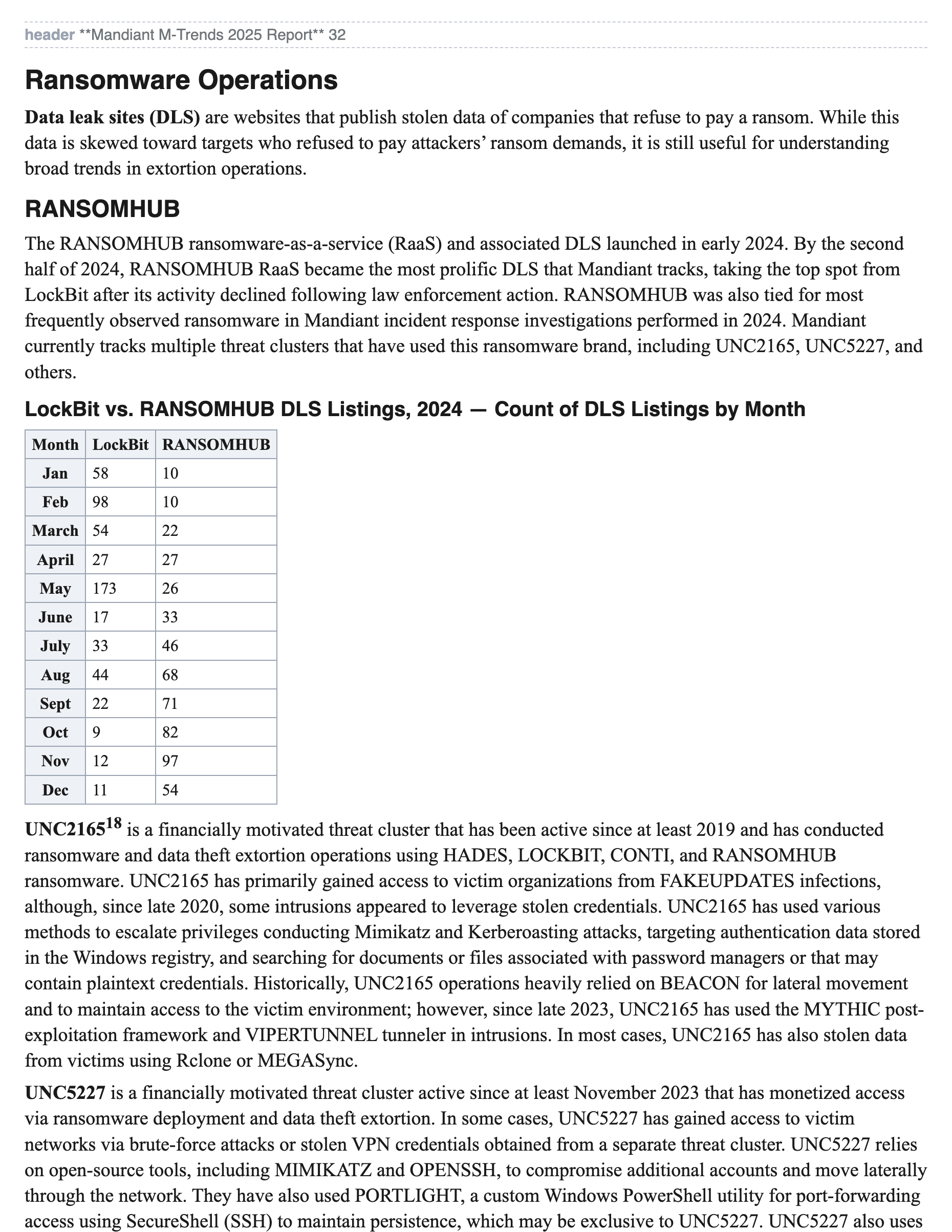}}\\[3pt]
{\scriptsize\sffamily\color{LIMuted}\ocria{} output, rendered}
\end{minipage}
\caption{\textbf{A chart without printed values.} A bar chart of monthly listings for two ransomware groups, with no value printed on the bars. \ocria{} transcribes it as a table of all 24 bars, reading each value from the axis. On six of the ten points that ParseBench tests, the reading is one or two units from the expected value, for example 58 against 60 and 98 against 100 for LockBit in January and February, and the benchmark counts each of these as a miss.}
\label{fig:limit-chart}
\end{figure}

\Cref{fig:limit-long-table,fig:limit-chart} show two pages at the limits of the current model.

\section{Serving stack}
\label{app:serving}

This appendix describes the serving stack of \Cref{inference}, how its speed is measured (\Cref{app:speed-protocol}) and the draft model (\Cref{app:speculator}).

The served model is the bf16 checkpoint quantized to a mixed NVFP4 and FP8 map~\citep{nvidia2025nvfp4}, calibrated on the model's own transcriptions and chosen by a per-layer search that kept only maps within the noise of the bf16 model on ParseBench ($-0.0039$, 95\,\% interval $[-0.0083, +0.0006]$ on 1,578 documents), with the vision encoder left in bf16.
It runs on vLLM~\citep{kwon2023vllm} with the draft of \Cref{speculator} and five changes:
\begin{itemize}
\item a port of ReplaySSM~\citep{vllm2026replayssm}, which keeps one checkpoint of each gated-delta-network state and replays the accepted tokens instead of snapshotting the state at every draft position, with its verify kernel rewritten in CUDA;
\item fused activation-quantization kernels;
\item the batch metadata built once per step instead of once per cache group;
\item the prompt prefix shared by every request, captured once and injected per request;
\item a draft length that switches from 31 tokens to 15 above thirteen concurrent requests, where the verify step becomes compute-bound.
\end{itemize}
Every change was kept only if its kernels produced the same bits as the ones they replaced or its ParseBench score on 749 documents stayed within the noise of the previous served model.

\subsection{Speed measurements}
\label{app:speed-protocol}

\paragraph{Concurrency sweep.}
Requests come from a client in the same region that keeps a fixed number of them in flight; the model decodes greedily under the production prompt on pages rendered at 4.1\,megapixels (about 3,900 image tokens), and per-request decoding speed is a request's output tokens divided by its own decoding time, as reported by AIPerf, the load generator we use.
With a single request, the model decodes at 4,734 tokens per second, and the median request of about 1,300 output tokens takes 0.41\,s from request to last token, time to first token included, about nine times the end-to-end speed of the bf16 checkpoint on stock vLLM (321 to 333 tokens per second on one page).
With eight concurrent requests, the model decodes at 1,932 tokens per second per request on production pages; most of the difference from the single-request figure is the prefill for the other requests, which runs on the same GPU and stalls the decoding steps it overlaps.
With 64 concurrent requests, the GPU parses 10.21 pages of production traffic per second with a p90 latency of 9.6\,s (11.78 with every page rescaled to 1,200 output tokens), and \Cref{tab:speed-results} gives the levels in between.
\Cref{fig:decode-ladder} shows the single-stream speed.

\begin{table}[!htbp]
\RCsetup
\begin{tabularx}{\linewidth}{>{\centering\arraybackslash}X>{\centering\arraybackslash}X>{\centering\arraybackslash}X>{\centering\arraybackslash}X}
\toprule
\RChead{Concurrent requests} & \RChead{Output tok/s per request} & \RChead{Aggregate output tok/s} & \RChead{Pages/s} \\
\midrule\RCstart
1 & 4,734 & -- & 2.01 \RCnext
8 & 1,932 & 9,964 & 7.42 \RCnext
16 & 1,103 & 12,740 & 8.67 \RCnext
32 & 566 & 14,833 & 9.77 \RCnext
48 & 400 & 15,240 & 9.98 \RCnext
64 & 302 & 15,557 & 10.21 \RClast
\bottomrule
\end{tabularx}
\caption{Speed of \ocria{} on one B200 with the final serving stack of \Cref{app:serving}, against the number of concurrent requests. Measured with AIPerf, each value is the mean of three runs. Aggregate output tokens per second is all output tokens of a run over the run's wall-clock time. \Cref{app:speed-protocol} gives the protocol.}
\label{tab:speed-results}
\end{table}

\begin{figure}[!htbp]
\centering
\begin{tikzpicture}
\begin{axis}[
  width=0.88\linewidth, height=4.0cm,
  xbar, bar width=11pt, bar shift=0pt,
  xmin=0, xmax=5600, xtick={0,1000,...,5000},
  xticklabel style={/pgf/number format/.cd, fixed, 1000 sep={{,}}},
  xlabel={tokens per second, single stream ($c = 1$), one B200},
  symbolic y coords={optimized, stock}, ytick={optimized, stock}, enlarge y limits=0.55,
  yticklabels={{\ocria{}, optimized stack}, {bf16, stock vLLM}},
  axis line style={LIPanelLine, line width=0.4pt},
  tick style={LIPanelLine, line width=0.4pt},
  xmajorgrids, major grid style={LIPanelLine, line width=0.3pt},
  tick label style={font=\scriptsize\sffamily, text=LIMuted},
  yticklabel style={font=\scriptsize\sffamily, text=LIInk, align=right},
  label style={font=\scriptsize\sffamily, text=LIInk},
  clip=false,
]
\addplot[xbar, fill=LIMuted, draw=none] coordinates {(327,stock)};
\addplot[xbar, fill=LIBlue,  draw=none] coordinates {(3000,optimized)};
\addplot[draw=none, mark=none, error bars/.cd, x dir=both, x explicit,
  error bar style={LIInk, line width=0.6pt}, error mark options={LIInk, rotate=90, mark size=2.2pt, line width=0.6pt}]
  coordinates {(327,stock) -= (6,0) += (6,0)};
\addplot[draw=none, mark=none, error bars/.cd, x dir=both, x explicit,
  error bar style={LIInk, line width=0.6pt}, error mark options={LIInk, rotate=90, mark size=2.2pt, line width=0.6pt}]
  coordinates {(3000,optimized) -= (1550,0) += (900,0)};
\addplot[only marks, mark=diamond*, mark size=3pt, LIInk] coordinates {(4734,optimized)};
\node[font=\scriptsize\sffamily, text=LIInk, anchor=west, inner sep=1.5pt, xshift=4pt] at (axis cs:333,stock) {321--333};
\node[font=\scriptsize\sffamily, text=LIInk, anchor=south, inner sep=1.5pt, yshift=7pt] at (axis cs:2675,optimized)
  {$\approx$ 3,000 end to end, median request; 1,450--3,900 browser demo};
\node[font=\scriptsize\sffamily, text=LIInk, anchor=north, inner sep=1.5pt, yshift=-7pt] at (axis cs:4734,optimized)
  {4,734 (AIPerf $c=1$, TTFT excluded)};
\end{axis}
\end{tikzpicture}
\caption{\textbf{Single-stream decoding speed on one B200.} The same model in bf16 on stock vLLM against the optimized stack (NVFP4 target, DFlash2 draft with 31 draft tokens, the kernels of \Cref{app:serving}), greedy decoding, the production prompt, one page at a time. Bars are the end-to-end rate, time to first token included, of the median request at a client in the same region; the whisker is the range seen from a browser client over runs; the diamond is the AIPerf per-request decode rate on the concurrency ladder, which excludes the time to first token.}
\label{fig:decode-ladder}
\end{figure}

\paragraph{Cross-model comparison.}
We compare \ocria{} with Nemotron-3.5-Parse, HunyuanOCR-1.5 with DFlash, SmolDocling-256M and Falcon-OCR in BF16, each on one B200 at one and eight concurrent requests.
\Cref{tab:speed-systems} gives the values plotted in \Cref{fig:speed-systems}.
\begin{table}[!htbp]
\RCsetup
\begin{tabularx}{\linewidth}{>{\raggedright\arraybackslash}Xccc}
\toprule
\RChead{Model} & \RChead{Concurrent requests} & \RChead{Output tok/s per request} & \RChead{Pages/s} \\
\midrule\RCstart
\ocria{} (optimized) & 1 & 4,734 & 2.01 \RCnext
Nemotron-3.5-Parse & 1 & 1,395 & 0.80 \RCnext
HunyuanOCR-1.5 + DFlash & 1 & 1,403 & 0.71 \RCnext
SmolDocling-256M & 1 & 894 & 0.55 \RCnext
Falcon-OCR bf16 & 1 & 599 & 0.39 \RClast
\midrule\RCstart
\ocria{} (optimized) & 8 & 1,932 & 7.83 \RCnext
Nemotron-3.5-Parse & 8 & 963 & 4.75 \RCnext
HunyuanOCR-1.5 + DFlash & 8 & 787 & 3.26 \RCnext
SmolDocling-256M & 8 & 732 & 3.21 \RCnext
Falcon-OCR bf16 & 8 & 356 & 1.95 \RClast
\bottomrule
\end{tabularx}
\caption{Cross-model speed comparison on one B200 with one and eight concurrent requests. \Cref{tab:speed-results} reports the separate concurrency sweep.}
\label{tab:speed-systems}
\end{table}

\FloatBarrier

\subsection{Speculator}
\label{app:speculator}

This appendix gives the details of the draft model of \Cref{speculator}.
Like other recent drafts~\citep{cai2024medusa,li2025eagle3,gloeckle2024mtp} it reads the hidden states the target computed for the context rather than the context's tokens: DFlash concatenates them from eight of the target's layers, projects them and injects them as precomputed keys and values into every draft layer, and the draft writes a whole block of tokens in one pass, with mask tokens standing in for the block positions.
The convolution lets each block position see the ones before it, and the selector keeps the sixteen most likely tokens at each position, scores every pair of consecutive candidates with a low-rank bilinear term, and walks one path through the block.

\paragraph{Training details.}
The training features are the target's hidden states at the eight layers, captured once by running the quantized target of the time over training pages with greedy decoding and the image-only transcription prompt; the target stays frozen, and the served draft was trained against the checkpoint that preceded the released one.
Training has two phases and then the selector.
First only the projection, its normalisation and the convolutions are trained, 72\,M of the 686\,M parameters, with the cross-entropy of the target's greedy tokens weighted by $\exp(-\mathrm{position}/4)$, because the early positions decide whether a step accepts anything at all; on 16,384 pages this takes about ten hours on one B200 and raises the accepted length on held-out replay from 8.2 to 13.0 tokens per block.
Then the whole draft is unfrozen and trained on 15,481 pages (18,142 steps on eight B200s, about four hours), warm-starting through 2k- and 8k-page stages; at equal data the unfrozen draft beats the frozen one (10.15 against 9.93 accepted tokens).
The selector is fitted last on the frozen draft from its cached features, with a prefix cross-entropy over the sixteen candidates at 32 anchor positions per page, so the quantity optimized is the length of the prefix the target will accept in one verification step; it raises the held-out accepted length from 12.65 to 14.87 tokens, and more pages helped at every stage (10.2, 11.2 and 12.7 accepted tokens from 1k, 4k and 16k pages).

\paragraph{Path selection.}
We refine DFlash~2's draft-token selection with dynamic programming.
DFlash~2 greedily selects among sixteen candidates per position using scores conditioned on the previous choice.
We normalise these scores into transition probabilities $P_i(c_i \mid c_{i-1})$, the probability that the target accepts candidate $c_i$ at position $i$ after $c_{i-1}$, and select the chain $c_1,\dots,c_K$ that maximises the expected accepted-prefix length:
\begin{equation}
\mathbb{E}[\text{accepted length}] = \sum_{l=1}^{K} \prod_{i \le l} P_i(c_i \mid c_{i-1}),
\end{equation}
where $c_0$ is the last verified token.
Solving this objective exactly adds negligible cost and gives slightly more accepted tokens per step than the default greedy selection (\Cref{fig:acceptance-positions}).

\paragraph{Acceptance.}
Served with 31 draft tokens per step, each verification step emits 13 to 17 tokens, that is 12 to 16 accepted drafts plus the target's own token, depending on the pages: 17.5 on the eight pages the kernel work was tuned on, 14.2 on the fixed set of pages run at every server start, and 13.3 on production traffic (\Cref{tab:acceptance}); with 15 draft tokens the same draft emits 12.2 on the tuning pages.
Published drafts with a 16-token block accept fewer: DFlash reports about 6.5 tokens per step on Qwen3-8B, the draft shipped with HunyuanOCR-1.5 about 8 to 9, and multi-token prediction heads about 5~\citep{dflash2026,hunyuanocr2026,duan2026glmocrtechnicalreport}.
Acceptance does not depend on the batch size, so the draft weights are the same at every batch size from 1 to 64; only the number of draft tokens changes, from 31 to 15 above thirteen concurrent requests (\Cref{app:serving}).
\Cref{fig:acceptance-positions} gives the acceptance by draft position: the probability that a draft token is right, given that the ones before it were, stays above 0.8 to the last position, so a block of 32 is not too long.

\begin{table}[!tbp]
\RCsetup
\begin{tabularx}{\linewidth}{>{\hsize=1.1\hsize}Xccc>{\hsize=0.9\hsize\raggedright\arraybackslash}X}
\toprule
\RChead{Setting} & \RChead{k} & \RChead{Tokens per step} & \RChead{Accepted per step} & \RChead{Condition} \\
\midrule
\RCgroup{\ocria{}, DFlash~2 draft, greedy}{tokens emitted per verification step; accepted drafts of 31}
Tuning pages, c\,=\,8 & 31 & 17.5 & 16.5 & timed inside the server, B200 \RCnext
Production pages, 40-page sample, c\,=\,1 & 31 & 12.5 & 11.5 & timed inside the server, B200 \RCnext
Production traffic, 512 pages, c\,=\,8 & 31 & 13.3 & -- & per-request mean, AIPerf ladder of \Cref{tab:speed-results} \RCnext
Fixed canary pages at container start, c\,=\,1 & 31 & 14.2 & -- & B200 \RCnext
The same canary pages on a B300 & 31 & 12.9--13.2 & -- & B300 arithmetic differs from B200; 14.2 on B200 \RClast
\RCgroup{Published speculators}{authors' figures, each with its own definition of accepted length}
DFlash, Qwen3-8B, block 16~\citep{dflash2026} & 16 & 6.5 & -- & text benchmarks, temperature 0 \RCnext
DFlash, Qwen3-4B, block 16~\citep{dflash2026} & 16 & 6.5 & -- & text benchmarks, temperature 0 \RCnext
DFlash~2, Qwen3.8-27B, block 8~\citep{inco2026dflash2} & 8 & 4.1--5.5 & -- & vendor blog \RCnext
HunyuanOCR-1.5 with DFlash, block 16~\citep{hunyuanocr2026} & 16 & 8.4 & -- & OmniDocBench, vLLM, batch 1 (8.9 in Transformers) \RCnext
GLM-OCR, MTP, 10 tokens per step~\citep{duan2026glmocrtechnicalreport} & 10 & 5.2 & -- & authors' average \RCnext
HPD-Parsing, P-MTP~\citep{hpdparsing2026} & -- & 6.6 & -- & A800, vLLM \RCnext
GravityOCR, diffusion self-speculation~\citep{gravityocr2026} & -- & 9.7 & -- & tokens per forward pass, drafting and verification counted as separate passes, SGLang \RClast
\bottomrule
\end{tabularx}
\caption{Accepted tokens per verification step. Top group: the served DFlash~2 draft (6 layers, block 32, selector top-16) on the \ocria{} NVFP4 target at k\,=\,31, temperature 0; a step emits the accepted drafts plus one target token, so 17.5 tokens per step means 16.5 of 31 drafts accepted. Tuning pages are the eight pages the kernel work was tuned on; production pages are a 40-page production sample and the 512-page AIPerf ladder of \Cref{tab:speed-results}; the canary is a fixed page set run at every container start. Bottom group: published figures, each with its authors' definition of accepted length.}
\label{tab:acceptance}
\end{table}

\begin{figure}[t]
\centering
\begin{tikzpicture}
\begin{axis}[
  width=0.92\linewidth, height=5.6cm,
  xmin=0.5, xmax=31.5, xtick={1,5,10,15,20,25,31},
  ymin=0, ymax=1.02, ytick={0,0.2,...,1.0},
  yticklabel style={/pgf/number format/.cd, fixed, precision=1},
  xlabel={draft position $p$ ($k = 31$)}, ylabel={P(accepted $\geq p$)},
  axis line style={LIPanelLine, line width=0.4pt},
  tick style={LIPanelLine, line width=0.4pt},
  ymajorgrids, major grid style={LIPanelLine, line width=0.3pt},
  tick label style={font=\scriptsize\sffamily, text=LIMuted},
  label style={font=\scriptsize\sffamily, text=LIInk},
  legend style={font=\scriptsize\sffamily, draw=none, fill=none, text=LIInk, row sep=-2pt},
  legend cell align=left, legend pos=south west,
  clip=false, set layers,
]
\begin{pgfonlayer}{axis background}
  \fill[LIPanelGray] (axis cs:15.5,0) rectangle (axis cs:23.5,1.02);
\end{pgfonlayer}
\draw[LIPanelLine, line width=0.4pt] (axis cs:15.5,0) -- (axis cs:15.5,1.02) (axis cs:23.5,0) -- (axis cs:23.5,1.02);
\addplot[LIMuted, densely dashed, line width=0.9pt] coordinates {
  (1,0.987) (2,0.970) (3,0.948) (4,0.930) (5,0.899) (6,0.877) (7,0.851) (8,0.827) (9,0.802) (10,0.777)
  (11,0.757) (12,0.738) (13,0.711) (14,0.687) (15,0.667) (16,0.640) (17,0.614) (18,0.587) (19,0.554) (20,0.527)
  (21,0.505) (22,0.481) (23,0.454) (24,0.428) (25,0.402) (26,0.377) (27,0.345) (28,0.324) (29,0.307) (30,0.292)
  (31,0.280)};
\addlegendentry{oracle: truth inside the top-16 (upper bound)}
\addplot[LIOrange, dotted, line width=1.1pt] coordinates {
  (1,0.932) (2,0.941) (3,0.950) (4,0.942) (5,0.937) (6,0.948) (7,0.942) (8,0.942) (9,0.931) (10,0.935)
  (11,0.945) (12,0.943) (13,0.933) (14,0.944) (15,0.926) (16,0.898) (17,0.901) (18,0.887) (19,0.893) (20,0.896)
  (21,0.892) (22,0.888) (23,0.888) (24,0.915) (25,0.904) (26,0.888) (27,0.886) (28,0.886) (29,0.873) (30,0.851)
  (31,0.821)};
\addlegendentry{P(hit $\mid$ prefix correct), served walk}
\addplot[LIBlue, line width=1.3pt, mark=*, mark size=1.2pt] coordinates {
  (1,0.932) (2,0.877) (3,0.833) (4,0.785) (5,0.735) (6,0.697) (7,0.657) (8,0.619) (9,0.576) (10,0.539)
  (11,0.509) (12,0.480) (13,0.448) (14,0.423) (15,0.391) (16,0.352) (17,0.317) (18,0.281) (19,0.251) (20,0.225)
  (21,0.201) (22,0.178) (23,0.158) (24,0.145) (25,0.131) (26,0.116) (27,0.103) (28,0.091) (29,0.080) (30,0.068)
  (31,0.056)};
\addlegendentry{served selector walk: P(accepted $\geq p$)}
\node[font=\scriptsize\sffamily, text=LIInk, anchor=south, inner sep=1pt] at (axis cs:8,1.03) {positions 1--15: 9.50 tok/step};
\node[font=\scriptsize\sffamily, text=LIInk, anchor=south, inner sep=1pt] at (axis cs:19.5,1.03) {16--23: 1.96};
\node[font=\scriptsize\sffamily, text=LIInk, anchor=south, inner sep=1pt] at (axis cs:27.5,1.03) {24--31: 0.79};
\node[font=\scriptsize\sffamily, text=LIInk, anchor=east, inner sep=1pt, align=right] at (axis cs:30.8,0.20) {5.6\,\% of steps\\ accept all 31};
\end{axis}
\end{tikzpicture}
\caption{\textbf{Acceptance by draft position.} Probability that the target accepts at least the first $p$ draft tokens, for the greedy walk over the draft's candidates (the served dynamic-programming walk accepts slightly more) and for an oracle that picks the right token whenever it is among the draft's sixteen candidates, with the hit rate at each position given the prefix was right; offline replay of 9,193 greedy draft calls on 100 pages with 31 draft tokens. The area under the walk curve is the accepted tokens per step.}
\label{fig:acceptance-positions}
\end{figure}

\section{Prompts}
\label{app:prompts}

This is the system prompt for transcription.

\begin{LIPromptBox}{Plain transcription prompt}
Transcribe this Page in markdown, be as accurate as possible in your transcription, do not describe/summarize.

Formatting:
- Use ** for bold text and * for italic text.
- Use ~~ text ~~ for strikethrough text. Only use this for text that is visually stricken through.
- Use <mark> and </mark> tags for highlighted text, aka text to wich a stabilo hilight was applied Only use to hilight sentence or words, not full block of text/tables. Use HTML color names for different colors of highlight, e.g. <mark style="background-color: blue">text</mark>. Do not use mark for first cell of a table, as it is often used for the header and should not be highlighted.
- Use <u> and </u> tags for underlined text.
- Use \* for bullet points and - for sub-bullets.
- Use > for blockquotes.
- Use ` for inline code.
- Use ``` for code blocks.
- Use [ ] for empty checkboxes.
- Use [x] for checked checkboxes.
- Use <sub> and </sub> for subscript text.
- Use <sup> and </sup> for superscript text.
- Use [___] for empty fields to fill, if the field is already filled, represent them as field content in markdown.
- Use $...$ for inline equations and $$...$$ for display equations. Do NOT use \(...\) or \[...\] — the corpus uses dollar delimiters (43,604 inline occurrences against 35 in the parenthesis form).
- Chemical formulas are NOT LaTeX: write them as plain text with <sub> and <sup>, e.g. K<sub>3</sub>Fe(CN)<sub>6</sub> and 2H<sub>2</sub>O, never $K_3Fe(CN)_6$. Reserve $...$ for real mathematics.

Tables:
- Output tables in HTML when possible.
- Table should not be use for page formatting (column or grid like layout), but only for tabular data.
- Some table may be irregular, with merged cells, missing cells, or inconsistent formatting, or have an exotic structure (like a triangular table). Do your best to represent the table accurately with correct reading order.

Charts, Graphs, and Figures:
- IMPORTANT: Every chart or graph MUST be converted into an HTML <table>. This is non-negotiable. The table represent the chart data, not the chart layout. Focus on accurately capturing the underlying data and relationships in the table format.
- FORBIDDEN outputs for charts: <iframe>, <embed>, <canvas>, <img>, <object>, links, URLs, or plain-text descriptions. Using any of these for a chart is an automatic FAIL.
- MERMAID FALLBACK: For diagrams, figures and flowcharts that are better represented as a graph than as a table (architecture diagrams, flowcharts, org charts, entity relationships), use Mermaid syntax INSTEAD of an HTML table and output it between ```mermaid and ``` tags. This exception applies ONLY to diagrams; a chart or graph plotting data values is still an HTML <table>.

Images:
- Transcribe any text printed inside the image as markdown.
- Reference the image itself as ![label: description], where label is exactly one of: logo, icon, photo, illustration, screenshot, map, signature, seal, code, drawing, image. Use the caption or wording the page itself prints for the description; do not invent one, and do not describe an image the page says nothing about.

Page furniture:
- Output the page header (if any) between <page_header> and </page_header> tags.
- Output the page footer (if any) between <page_footer> and </page_footer> tags.
- Output the printed page number (if any) between <page_number> and </page_number> tags, nested inside the header or footer band it sits in.

Watermarks:
- Output a watermark stamped across the page (DRAFT, OBSOLETE, CONFIDENTIAL) between <watermark> and </watermark> tags, not as body text.

Multi-column Text:
- Unroll multi-column text into a single column format.

Completeness:
- Ensure ALL content from the page is transcribed without omissions.
- As this is a transcription task, fidelity is of upmost importance. You should NEVER truncate or shorten tabular data. Any instance of table data that looks like "[...table data continues in rows n - m ...]" or omits rows or columns will result in an automatic FAIL.

Accuracy:
- Maintain text and formatting as close to the original as possible within the constraints of Markdown syntax.

Handwriting / Cursive Text:
- If the page contains handwriting or cursive text, ignore the OCR output for those sections and rely solely on the screenshot to transcribe that content.

Output result between ```markdown and ```
\end{LIPromptBox}

\clearpage
\section{Examples}
\label{app:examples}

This appendix shows \ocria{} on pages from ParseBench~\citep{zhang2026parsebenchdocumentparsingbenchmark}.
We chose the pages to show one capability each, so they are not a sample of the benchmark; \Cref{evaluations} gives the scores over all pages.
Each page is rendered at 150\,dpi and transcribed with the prompts of \Cref{app:prompts} and greedy decoding.
The right-hand panel of each figure renders what the model wrote, without editing: tables as the HTML the model produced, page headers and footers in grey, image markers in brackets, and formulas typeset from the model's \LaTeX.
Where an output is much longer than the page image, the panel shows its beginning.

\begin{figure}[!htbp]
\centering\setlength{\fboxsep}{0pt}%
\begin{minipage}[t]{0.49\textwidth}\vspace{0pt}\centering
\fcolorbox{LIPanelLine}{white}{\includegraphics[width=\dimexpr\linewidth-0.8pt\relax]{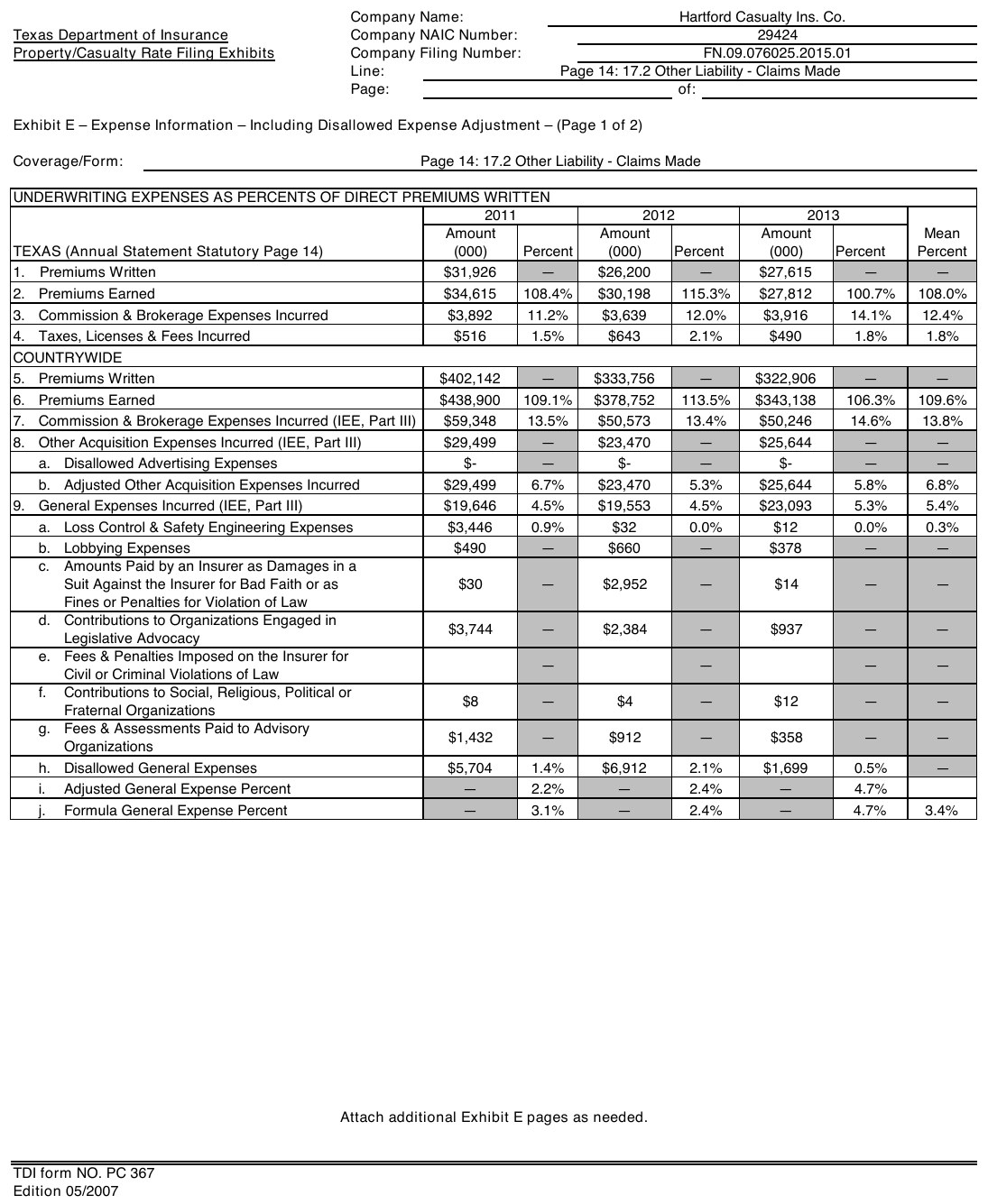}}\\[3pt]
{\scriptsize\sffamily\color{LIMuted}Page image}
\end{minipage}\hfill
\begin{minipage}[t]{0.49\textwidth}\vspace{0pt}\centering
\fcolorbox{LIPanelLine}{white}{\includegraphics[width=\dimexpr\linewidth-0.8pt\relax]{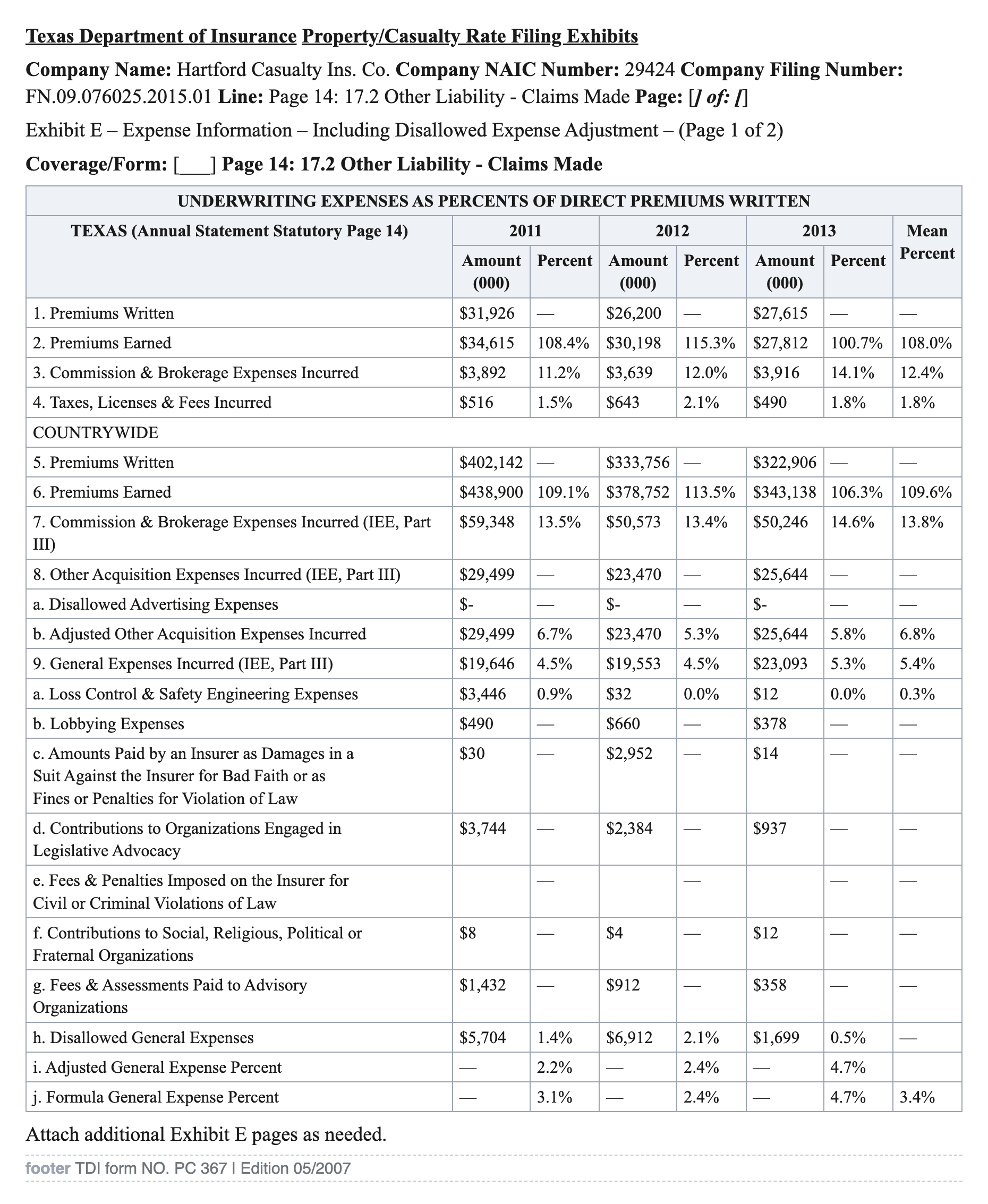}}\\[3pt]
{\scriptsize\sffamily\color{LIMuted}\ocria{} output, rendered}
\end{minipage}
\caption{\textbf{A dense financial table.} A page of a Texas insurance rate filing, with a three-level header: a title row over the years 2011 to 2013, and an amount and a percent column under each year. \ocria{} writes the page as one HTML table, with the header spans as \LIcode{colspan} attributes and the section rows (\textit{COUNTRYWIDE}) kept in place.}
\label{fig:ex-table}
\end{figure}

\begin{figure}[!htbp]
\centering\setlength{\fboxsep}{0pt}%
\begin{minipage}[t]{0.49\textwidth}\vspace{0pt}\centering
\fcolorbox{LIPanelLine}{white}{\includegraphics[width=\dimexpr\linewidth-0.8pt\relax]{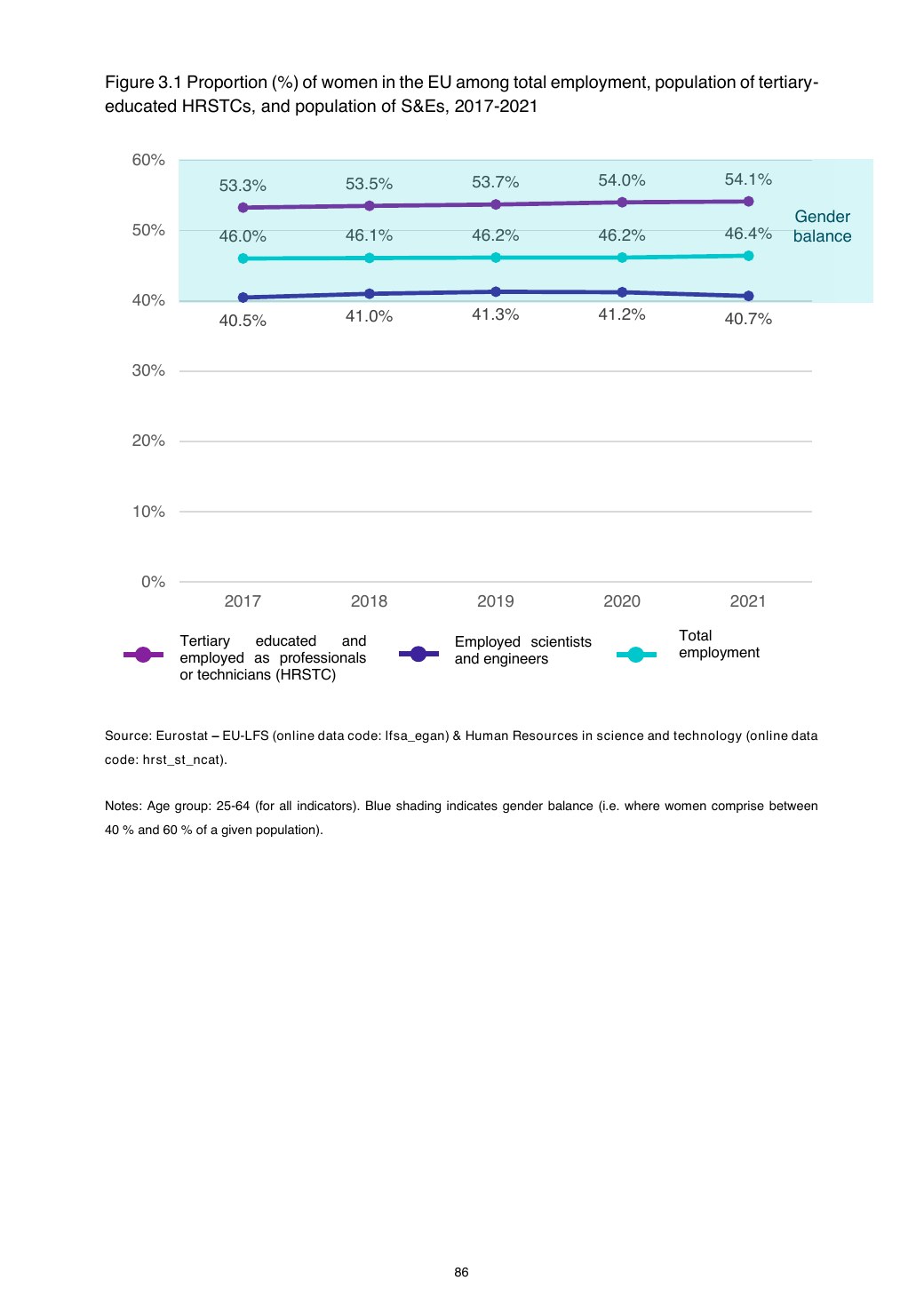}}\\[3pt]
{\scriptsize\sffamily\color{LIMuted}Page image}
\end{minipage}\hfill
\begin{minipage}[t]{0.49\textwidth}\vspace{0pt}\centering
\fcolorbox{LIPanelLine}{white}{\includegraphics[width=\dimexpr\linewidth-0.8pt\relax]{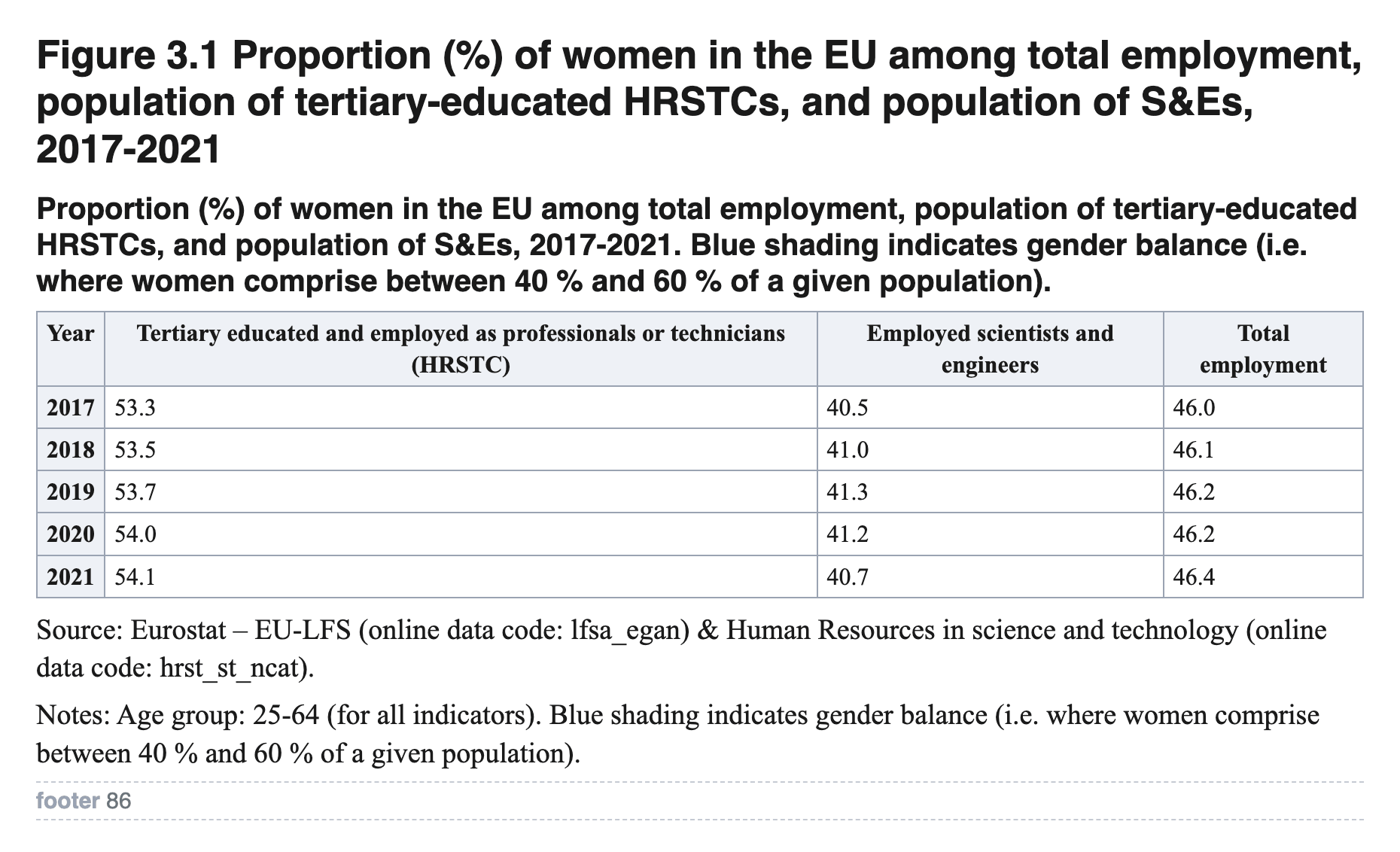}}\\[3pt]
{\scriptsize\sffamily\color{LIMuted}\ocria{} output, rendered}
\end{minipage}
\caption{\textbf{A chart transcribed as its data.} A line chart with three series over five years. \ocria{} writes the chart as a table of its 15 data points, one row per year and one column per series, and the values match the labels printed on the chart.}
\label{fig:ex-chart}
\end{figure}

\begin{figure}[!htbp]
\centering\setlength{\fboxsep}{0pt}%
\begin{minipage}[t]{0.49\textwidth}\vspace{0pt}\centering
\fcolorbox{LIPanelLine}{white}{\includegraphics[width=\dimexpr\linewidth-0.8pt\relax]{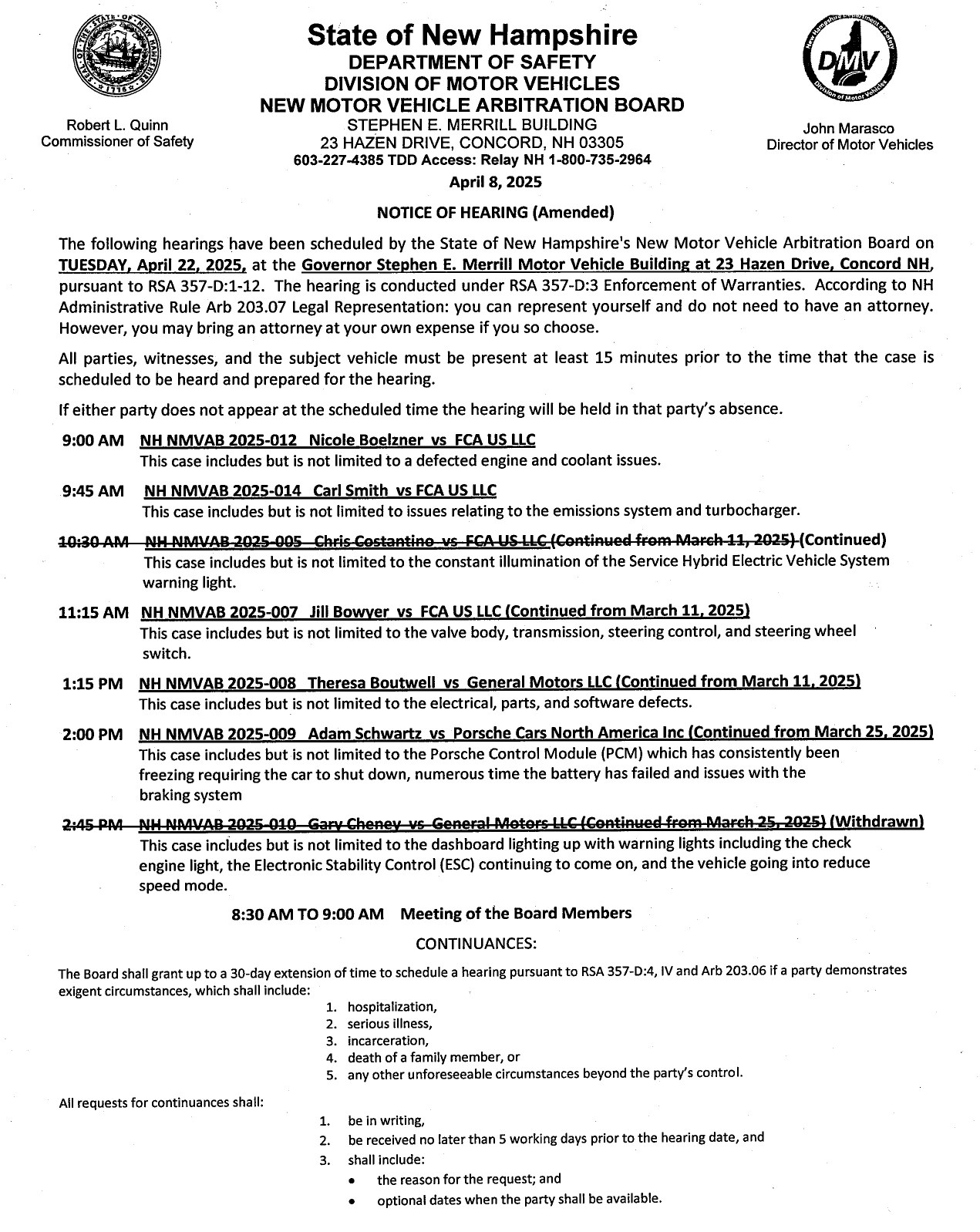}}\\[3pt]
{\scriptsize\sffamily\color{LIMuted}Page image}
\end{minipage}\hfill
\begin{minipage}[t]{0.49\textwidth}\vspace{0pt}\centering
\fcolorbox{LIPanelLine}{white}{\includegraphics[width=\dimexpr\linewidth-0.8pt\relax]{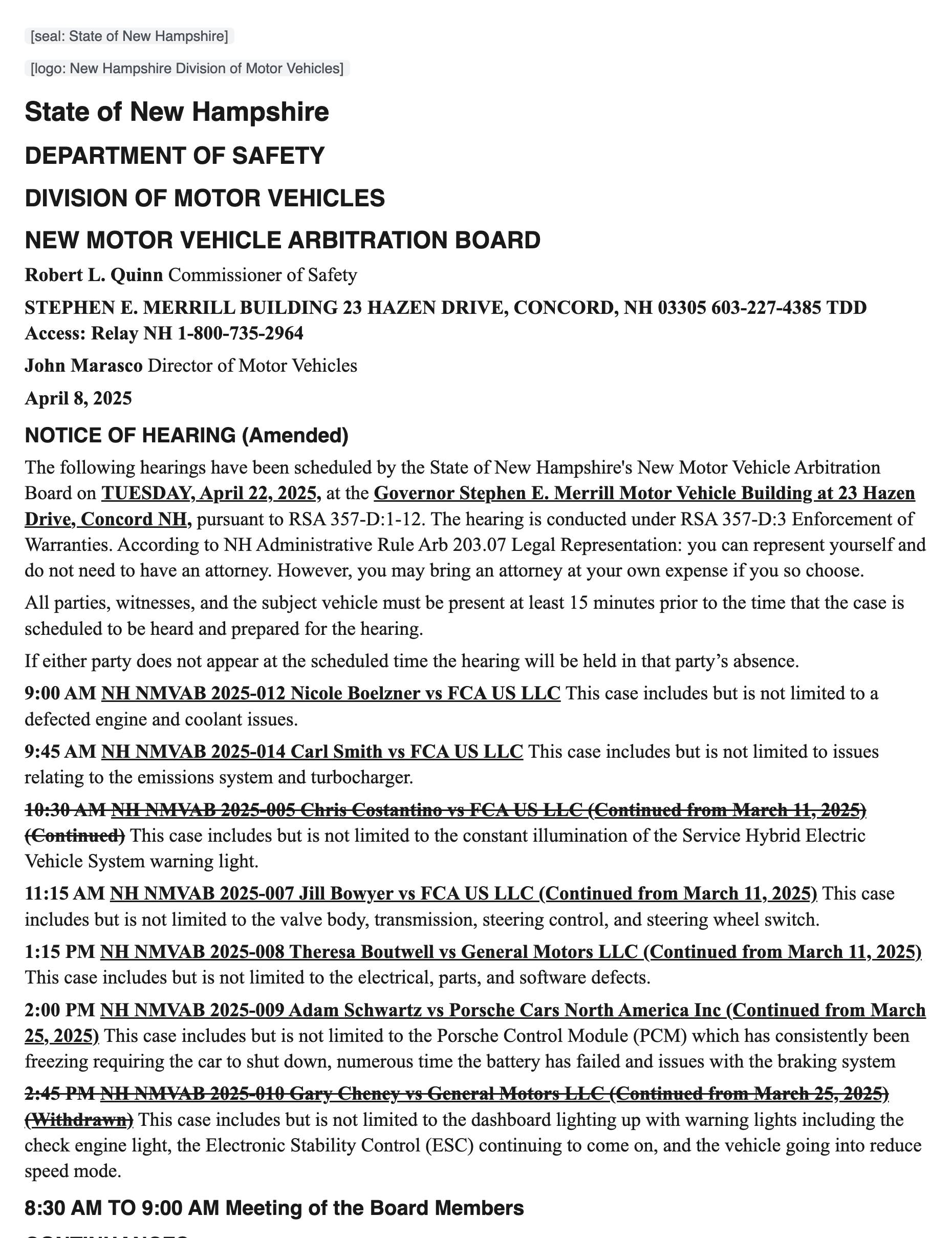}}\\[3pt]
{\scriptsize\sffamily\color{LIMuted}\ocria{} output, rendered (beginning)}
\end{minipage}
\caption{\textbf{Text formatting.} A hearing notice that marks a withdrawn and a continued case by striking them through, underlines the case names and sets the times in bold. \ocria{} keeps all three: the struck-through entries as Markdown strikethrough, the case names as \LIcode{<u>} and the times as bold.}
\label{fig:ex-format}
\end{figure}

\begin{figure}[!htbp]
\centering\setlength{\fboxsep}{0pt}%
\begin{minipage}[t]{0.49\textwidth}\vspace{0pt}\centering
\fcolorbox{LIPanelLine}{white}{\includegraphics[width=\dimexpr\linewidth-0.8pt\relax]{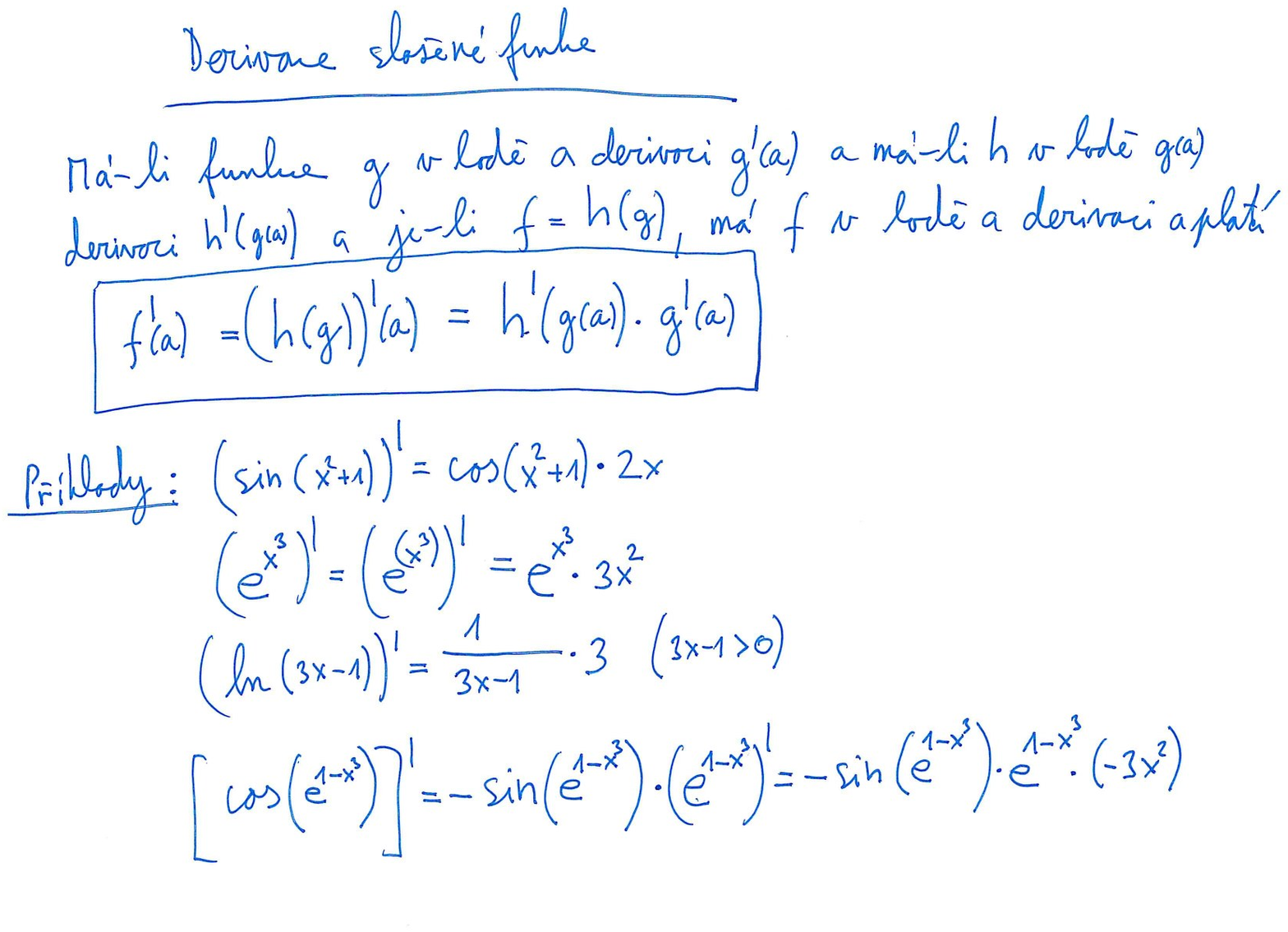}}\\[3pt]
{\scriptsize\sffamily\color{LIMuted}Page image}
\end{minipage}\hfill
\begin{minipage}[t]{0.49\textwidth}\vspace{0pt}\centering
\fcolorbox{LIPanelLine}{white}{\includegraphics[width=\dimexpr\linewidth-0.8pt\relax]{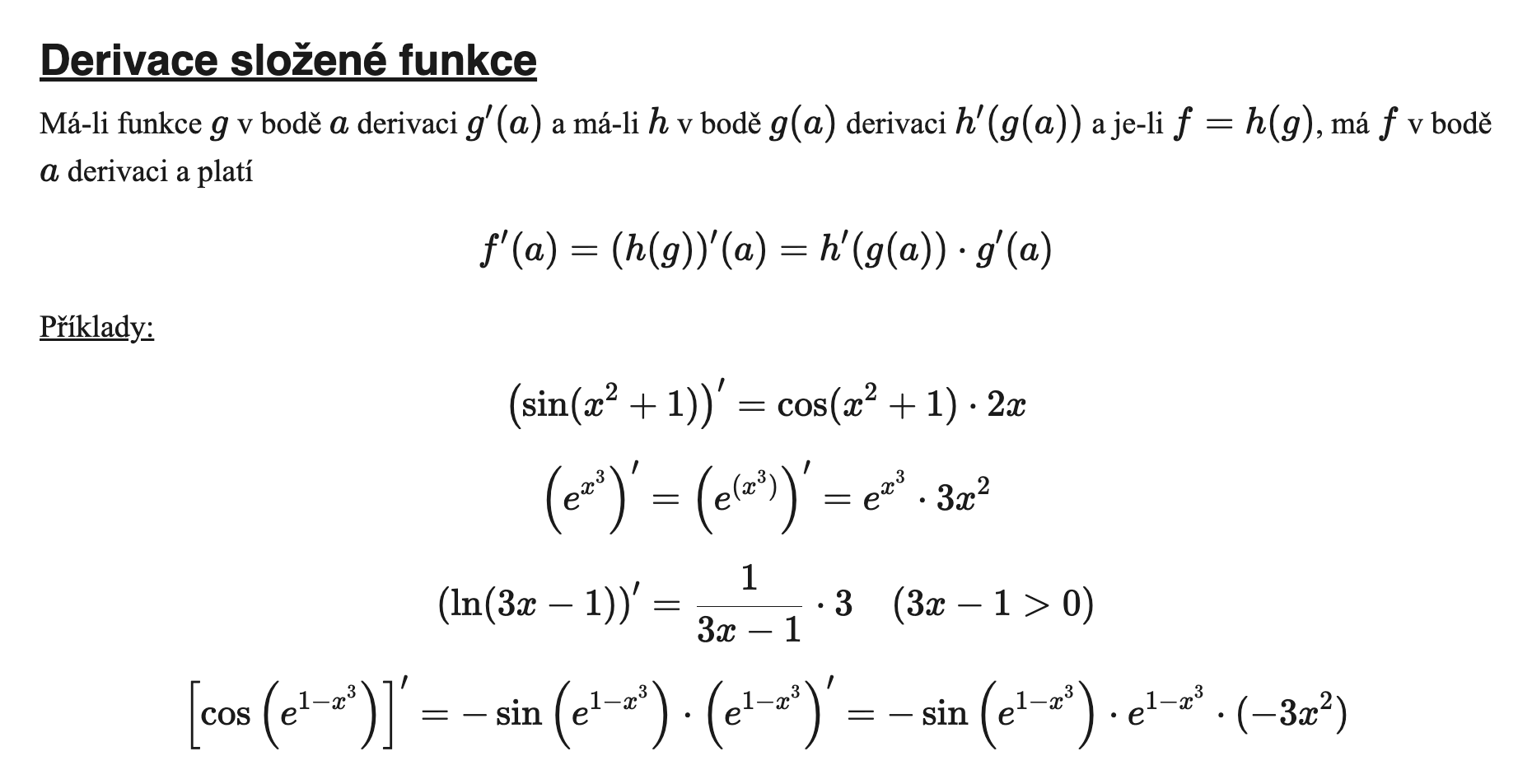}}\\[3pt]
{\scriptsize\sffamily\color{LIMuted}\ocria{} output, rendered}
\end{minipage}
\caption{\textbf{Handwritten formulas.} A handwritten note in Czech on the derivative of a composite function. \ocria{} transcribes the Czech text and writes each formula as \LaTeX.}
\label{fig:ex-handwriting}
\end{figure}

\subsection*{Two pages at the limits of the current model}

The two pages below show where \ocria{} reaches the limits of the current model.
In both, the model gets the structure of the page right, and the error is in the last part of the work: the rows past the output budget, and the exact values of a chart that prints none.

\end{document}